%% file: main.tex
\pdfoutput=1
\documentclass{article} 
\usepackage{iclr2022_conference,times}

\input{math_commands.tex}

\usepackage{hyperref}
\usepackage{url}

\usepackage{makecell}
\usepackage{stfloats}
\usepackage{amsmath}
\usepackage{amsbsy}
\usepackage{amssymb}
\usepackage{xcolor}
\usepackage{graphicx}
\usepackage[export]{adjustbox}
\usepackage{subcaption}
\usepackage{footnote}
\usepackage{multirow}
\makesavenoteenv{tabular}
\makesavenoteenv{table}
\usepackage{url}
\usepackage{tikz}
\usetikzlibrary{fit,positioning}
\usepackage{booktabs}
\usepackage{tabularx}
\usepackage{makecell}
\usepackage{tikz-dependency}
\usetikzlibrary{decorations}
\usetikzlibrary{decorations.pathreplacing}
\usepackage[ruled,vlined]{algorithm2e}

\usepackage[english]{babel} 

\expandafter\def\expandafter\UrlBreaks\expandafter{\UrlBreaks
  \do\a\do\b\do\c\do\d\do\e\do\f\do\g\do\h\do\i\do\j%
  \do\k\do\l\do\m\do\n\do\o\do\p\do\q\do\r\do\s\do\t%
  \do\u\do\v\do\w\do\x\do\y\do\z\do\A\do\B\do\C\do\D%
  \do\E\do\F\do\G\do\H\do\I\do\J\do\K\do\L\do\M\do\N%
  \do\O\do\P\do\Q\do\R\do\S\do\T\do\U\do\V\do\W\do\X%
  \do\Y\do\Z}

\usepackage{soulutf8}
\usepackage[normalem]{ulem}
\usepackage{enumitem}
\setlist{nolistsep}
\usepackage{microtype}

\input{draft_final}

\DeclareMathOperator*{\Gammaopenc}{\Gamma^{enc}}

\DeclareMathOperator*{\Gammaopdec}{\Gamma^{dec}}
\DeclareMathOperator*{\argr}{arg_{\textit r}}
\DeclareMathOperator*{\rop}{\rho_{\textit r}}

\title{Towards Unsupervised Content \\ Disentanglement in Sentence Representations\\ via Syntactic Roles}

\author{Ghazi Felhi
, Joseph Le Roux
LIPN\\
Université Sorbonne Paris Nord-CNRS UMR 70301\\
Villetaneuse, France \\
\texttt{\{felhi, leroux\}@lipn.fr} \\
\And
Djamé Seddah \\
INRIA Paris\\
Paris, France \\
\texttt{djame.seddah@inria.fr} \\
}

\iclrfinalcopy 

\begin{document}

\maketitle

\begin{abstract}
  Linking neural representations to linguistic factors is crucial in order to build and analyze NLP models interpretable by humans. Among these factors, syntactic roles (e.g. subjects, direct objects,\dots)  and their realizations are essential markers since they can be understood as a decomposition of predicative structures and thus the meaning of sentences.
Starting from a deep probabilistic generative model with attention, we measure the interaction between latent variables and realizations of syntactic roles and show that it is possible to obtain, without supervision, representations of sentences where different syntactic roles correspond to clearly identified different latent variables. 
The probabilistic model we propose is an Attention-Driven Variational Autoencoder (ADVAE). Drawing inspiration from Transformer-based machine translation models, ADVAEs enable the analysis of the interactions between latent variables and input tokens through attention. We also develop an evaluation protocol to measure disentanglement with regard to the realizations of syntactic roles. This protocol is based on attention maxima for the encoder and on latent variable perturbations for the decoder. Our experiments on raw English text from the SNLI dataset show that \textit{i)} disentanglement of syntactic roles can be induced without supervision, \textit{ii)}  ADVAE separates syntactic roles better than classical sequence VAEs and Transformer VAEs, \textit{iii)} realizations of syntactic roles can be separately modified in sentences by mere intervention on the associated latent variables. Our work constitutes a first step towards unsupervised controllable content generation.
  The code for our work is publicly available\footnote{\href{https://github.com/ghazi-f/ADVAE}{https://github.com/ghazi-f/ADVAE}}.
\end{abstract}

\section{Introduction}

A disentangled representation of data describes information as a combination of separate \emph{understandable} factors. 
This separation provides better transparency, but also better transfer performance \citep{higgins2018darla, dittadi2021on}. When it comes to disentanglement, Variational Autoencoders \citep[VAEs;][]{DBLP:journals/corr/KingmaW13} were extensively proven effective \citep{Higgins2019-VAE:Framework, Chen2018c, Rolinek2019VariationalAccident}.
 and were used throughout several recent works 
\citep{Chen2019ARepresentations,Li2020ProgressiveRepresentations,John2020DisentangledTransfer}. 
  In NLP, disentanglement has been mostly performed to separate the semantics
(or content) in a sentence from characteristics such as style and structure in order to generate paraphrases \citep{Chen2019ARepresentations, John2020DisentangledTransfer, Bao2020, huang-chang-2021-generating, huang-etal-2021-disentangling}.
We show in our work that the information in the content itself can be separated with a VAE-based model. In contrast to the aforementioned works, we use neither supervision nor input syntactic information for this separation. We demonstrate this ability by controlling the lexical realization of core syntactic roles. For example, the subject in a sentence can be encoded separately and controlled to generate the same sentence with another subject. Our framework includes a model and an evaluation protocol aimed at measuring the disentanglement of syntactic roles.

The model we introduce is an Attention-Driven VAE (ADVAE), which we train  on the SNLI raw text dataset 
\citep{Schmidt2020AutoregressiveLoops}. It draws its inspiration from attention-based machine translation models \citep{Bahdanau2015NeuralTranslate,Luong2015EffectiveTranslation}.
 Such models translate sentences between languages with different underlying structures and can be inspected to show a coherent 
 alignment between spans from both languages. Our ADVAE uses Transformers \citep{Vaswani2017}, an attention-based architecture, to map sentences from
 a language to independent latent variables, then map these variables back to the same sentences. Although ADVAE could be used to study other attributes, we motivate it (\S \ref{MOTIVATION}) and therefore study it for the alignment of syntactic roles with latent variables.
  
Evaluating disentanglement with regard to spans is challenging. After training the model and only for evaluation, we use linguistic information (from an off-the-shelf dependency parser) to first extract
  syntactic roles from sentences, and then study their relation to latent variables. To study this relation on the
   ADVAE decoder, we repeatedly \textit{i)} generate a sentence from a sampled latent vector
  \textit{ii)} perturb this latent vector at a specific location
  \textit{iii)} generate a sentence from this new vector and observe the difference.
  On the encoder side, we study the attention values to see whether each latent variable is focused on a particular syntactic role 
  in input sentences. 
  The latter procedure is only possible through the way our ADVAE uses attention to produce latent variables. To the best of our knowledge, we are the first to use this transparency mechanism to obtain quantitative results for a latent variable model. 
  
  We first justify our focus on syntactic roles in \S \ref{SyntacticRoles}, then we go over our contribution, which is threefold: \textit{i)} We introduce the ADVAE, a model that is designed for \emph{unsupervised}
    disentanglement of syntactic roles, and that enables analyzing the interaction
    between latent variables and observations through the values of attention (\S\ref{PROPOSEDMODEL}), \textit{ii)} We design an experimental protocol for the challenging assessment of disentanglement over realizations of syntactic roles, based on perturbations on the decoder side and attention on the encoder side (\S\ref{EVALUATIONPROC}), \textit{iii)} Our empirical results show that our architecture disentangles syntactic roles better than standard sequence VAEs and Transformer VAEs and that it is capable of controlling realizations of syntactic roles separately during generation (\S\ref{EXPE}).

\section{Related Works}

\label{RELATED}
\paragraph{Linguistic information in neural models} 
Accounting for linguistic information provided better inductive bias in the design of neural NLP systems during recent years. For instance, successful attempts at capturing linguistic information with neural models helped improve grammar induction (RNNG; \citealp{Dyer2016RecurrentGrammars}),
 constituency parsing and language modeling (ON-LSTM; \citealp{Shen2019OrderedNetworks}, ONLSTM-SYD; \citealp{Du2020ExploitingApproach}), as well as controlled generation (SIVAE; \citealp{Zhang2020Syntax-infusedGeneration}). 
  Many ensuing works 
have also dived into the linguistic capabilities of the resulting models, the types of linguistic annotations that emerge best in them, and syntactic error
analyses \citep{Hu2020AModels, Kodner2020OverestimationModels, Marvin2020TargetedModels, Kulmizev2020DoFormalisms}. Based on the Transformer architecture, the self-supervised model BERT \citep{Devlin2018} has also been subject to studies showing that the linguistic information it captures is organized among its layers in a way remarkably close to the way a classical NLP pipeline
   works \citep{Tenney2020BERTPipeline}. Furthermore,
   \citep{Clark2019WhatAttentionb}, showed that many attention heads in BERT specialize in dependency parsing.
    We refer the reader to  \citep{Rogers2020AWorks} for an extensive review of Bert-related studies. However, 
   such studies most often rely on structural probes  \citep{Jawahar2019WhatLanguageACL2019, liu-etal-2019-linguistic, hewitt-manning-2019-structural} to explain representations, probes which are not without issues, as shown by  \citet{pimentel-etal-2020-information}.
In that regard, the generative capabilities and the attention mechanism of our model offer an alternative to probing: analysis is performed directly on sentences generated by the model and on internal attention values.

\paragraph{Disentanglement in NLP}
The main line of work in this area revolves around using multitask learning to 
separate concepts in neural representations (\textit{e.g.} style vs content \citep{John2020DisentangledTransfer}, syntax vs 
semantics \citep{Chen2019ARepresentations, Bao2020}). Alternatively,  \citet{huang-chang-2021-generating} 
and \citet{huang-etal-2021-disentangling} use syntactic trees \emph{as inputs} to separate syntax from semantics, and generate
paraphrases without a paraphrase corpus. 
Towards less supervision, \citet{Cheng2020ImprovingGuidance} only uses style information to separate style from content in 
representations. Literature on \emph{unsupervised} disentanglement in NLP remains sparse. Examples are 
the work of \citet{Xu2020OnSupervision} on categorical labels (sentiment and topic), and that of \citet{behjati2021inducing} on representing morphemes using character-level Seq2Seq models. The work of \citet{behjati2021inducing} is closest to ours as it uses Slot Attention \citep{locatello2020objectcentric}, which, like ADVAE, is a cross-attention-based representation technique. 
Our contribution depart from previous work since \emph{i)} syntactic parses are not used as learning signals but as a way to interpret our model, and \emph{ii)} cross-attention enables our model to link a fixed number of latent variables to text spans.

\section{Syntactic Roles and Dependency Parsing}
\label{SyntacticRoles}
\begin{figure*}[!h]
\centering
    \begin{minipage}[b]{0.48\linewidth}
    \begin{adjustbox}{minipage=\linewidth,scale=0.9}
        \begin{dependency}[theme = default]
           \begin{deptext}[column sep=1em, row sep=.1ex]
               DET  \&ADJ  \&NOUN  \&VERB  \&DET  \&ADJ  \&NOUN \\
              A \& talented \& musician \& holds \& his \& nice \& guitar \\
           \end{deptext}
           \deproot[edge style={orange!70!black},
            label style={orange!70!black, fill=white, text=orange!70!black}]{4}{ROOT}
           \depedge[edge style={blue},
            label style={blue, fill=white, text=blue}]{4}{3}{nsubj}
           \depedge[edge style={green!60!black},
            label style={green!60!black, fill=white, text=green!60!black}]{4}{7}{dobj}
           \depedge{3}{2}{amod}
           \depedge{3}{1}{det}
           \depedge{7}{6}{amod}
           \depedge{7}{5}{poss}
           \wordgroup[style={orange!70!black, fill=white}]{2}{4}{4}{pred}
           \wordgroup[style={blue, fill=white}]{2}{1}{3}{a0}
           \wordgroup[style={green!60!black, fill=white}]{2}{5}{7}{a1}
           \groupedge[edge style={blue},
            label style={blue, fill=white, text=blue}, edge below]{pred}{a0}{ARG0}{2ex} 
           \groupedge[edge style={green!60!black},
            label style={green!60!black, fill=white, text=green!60!black}, edge below]{pred}{a1}{ARG1}{2ex} 
        \end{dependency}
            \end{adjustbox}
    \end{minipage}
    \begin{minipage}[b]{0.48\linewidth}
    \begin{adjustbox}{minipage=\linewidth,scale=0.9}
        \begin{tikzpicture}
            \draw [color=white](0,0) rectangle (3,2);
            \draw[decorate,decoration={brace,raise=0.1cm}]
            (3,3.5) -- (3,1.7) node[align=center,above=0.2cm,pos=0.8, xshift=1.2cm] {Dependency \\Parse};
            
            \draw[decorate,decoration={brace,raise=0.1cm}]
            (3,1.65) -- (3,1.2) node[align=center,above=-0.2cm,pos=0.8, xshift=1.2cm] {PoS Tags};
            \draw[decorate,decoration={brace,raise=0.1cm}]
            (3,1.15) -- (3,0.0) node[align=center,above=-0.2cm,pos=0.8, xshift=1.2cm] {Predicative\\ Structure};
        \end{tikzpicture}
    \end{adjustbox}
    \end{minipage}

    \caption{A sentence and its syntactic roles. The correspondence between syntactic roles and elements of the predicative structure is highlighted with colors.}
    \label{fig:depanaex}
    
\end{figure*}
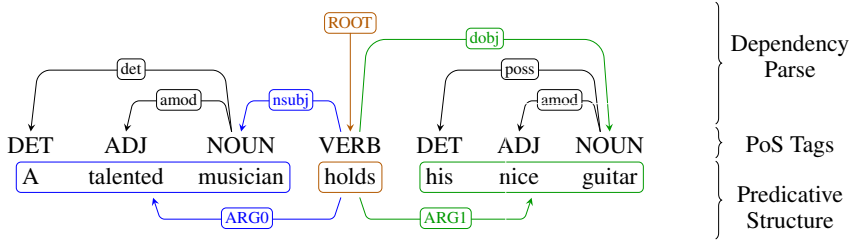
We present in Figure~\ref{fig:depanaex} an example sentence with its dependency parse\footnote{Following the ClearNLP constituent to dependency conversion, close to Stanford Dependencies \cite{de-marneffe-manning-2008-stanford}. See \url{https://github.com/clir/clearnlp-guidelines/blob/master/md/components/dependency_conversion.md}.}, its Part-of-Speech (PoS) tags, and a flat predicative structure with PropBank-like semantic roles \citep{palmer2005proposition}. Dependency parsing yields a tree, where edges are labeled with syntactic roles (or relations or functions) such as \emph{nominal subject} (\emph{nsubj}). The lexical realizations of these syntactic functions are textual spans and correspond to syntactic constituents. For instance, the lexical realization of the \emph{direct object} (\emph{dobj}) of the verb \emph{holds} in this sentence is the span \emph{his nice guitar}, with {\em guitar} as head. In short, the spans corresponding to subtrees consist of tokens that are more dependent of each other than of the rest of the sentence. As a consequence, and because a disentanglement model seeks independent substructures in the data, we expect such a model to converge to representations that display separation in realizations of frequent syntactic roles.  

In our work, we focus\footnote{Future research that takes interest in the finer-grained disentanglement of content may simply study a larger array of syntactic roles. Using our current system we display results including all syntactic roles in Appendix~\ref{ENTIREROLERESULTS}.} within the same framework. on nominal subjects, verbal roots of sentences, and direct or prepositional objects. These are \emph{core} (as opposed to \emph{oblique}; see \citet{nivre-etal-2016-universal} for details on the distinction) syntactic roles, since they directly relate to the predicative structure. In fact in most cases, as illustrated in Figure \ref{fig:depanaex}, the verbal root of a sentence is its main predicate, the nominal subject its agent (\emph{ARG0}) and the direct or prepositional object its patient (\emph{ARG1}). 


\section{Model Description}
\label{PROPOSEDMODEL}
The usual method to obtain sentence representations from Transformer models uses only a Transformer encoder either by taking an average of the token representations or by using the representation of a special token (\textit{e.g} [CLS] in BERT\citep{Devlin2018}). Recently, the usage of both Transformer encoders and decoders has also been explored in order to obtain representations whether by designing classical Autoencoders \citep{lewis2019bart, siddhant2019evaluating, raffel2020exploring},  or VAEs \citep{li2020optimus}. Our model, the ADVAE, differs from these models in that it uses \emph{targets} in Machine Translation (MT) Transformers (\textit{i.e} an encoder and a decoder) to produce sentence representations. 
Producing representations with Cross-Attention has been introduced by \cite{locatello2020objectcentric} as part of the Slot Attention modules in the context of unsupervised object discovery. However, in contrast to \cite{locatello2020objectcentric}, we simply use Cross-Attention as it is found in \cite{Vaswani2017}, \textit{i.e.} without normalizing attention weights over the query axis, or using GRUs\citep{cho2014learning} to update representations. As will be shown through our experiments, this is sufficient to disentangle syntactic roles. We explain the observation that motivates our work  
in \S\ref{MOTIVATION}, we then describe in \S\ref{ARCHITECTURE} the minimal changes we apply to MT Transformers, and finally,  we present the objective we use in \S\ref{OBJECTIVE}. The parallel between our model and MT Transformers is illustrated in Figure \ref{fig:SCHEMES}.

\begin{figure*}[!h]
\centering
    \begin{minipage}[b]{0.33\linewidth}
    \begin{adjustbox}{minipage=\linewidth,scale=0.53}
        \fbox{\begin{tikzpicture}
        \tikzstyle{main}=[rounded corners, minimum size = 10mm, thick, draw =black!80, node distance = 5mm, font=\Large, text centered]
        \tikzstyle{connect}=[-latex, thick]
        
        \node[main,text width=3cm] (source) [] {$s_1, s_2, ..., s_{N_s}$};
        \node[main,text width=2cm] (spe) [above=of source] {Positional Encoding};
        \node[main,text width=2.5cm] (enc) [above=of spe] {Transformer Encoder};
          
        \node[main,text width=4cm] (ptarget) [right=5 mm of source] {$t_1, t_2, ..., t_{N_t-1}$};
        \node[main,text width=2cm] (tpe) [above=of ptarget] {Positional Encoding};
        
        \node[minimum height = 5.8cm, minimum width = 5.5cm, thick, draw =black!80, node distance = 5mm, font=\Large, align=left, dashed, yshift=-0.2cm] (dec) [above=of tpe] {};
        \node[main,text width=5cm, align=left] (att) [above=of tpe] {Masked Self Attention + Cross-attention to align source information with target};
        \node[main,text width=5cm] (mlp) [above=of att] {MLP};
        \node[minimum width = 2.8cm, thick, draw =black!80, node distance = 5mm, font=\Large, align=left, minimum height = 1cm,  text width=2.8cm, xshift=-12mm, dashed] (declab) [above=of mlp] {Transformer\\ Decoder};
        
        \node[main,text width=5cm] (result) [above=20 mm of mlp] {$t_1, t_2, ..., t_{N_t}$};
        
       \path (source) edge [connect] (spe)
            (spe) edge [connect] (enc)
            (enc) edge [bend left][connect] (att)
            (ptarget) edge [connect] (tpe)
            (tpe) edge [connect] (att)
            (att) edge [connect] (mlp)
            (mlp) edge [bend right][connect] (result);
         \end{tikzpicture}}
    \end{adjustbox}
    \vskip -5px
    \caption*{(a) An MT Transformer}
    \end{minipage}
    \begin{minipage}[b]{0.33\linewidth}
    \begin{adjustbox}{minipage=\linewidth,scale=0.53}
        \fbox{\begin{tikzpicture}
        \tikzstyle{main}=[rounded corners, minimum size = 10mm, thick, draw =black!80, node distance = 5mm, font=\Large, text centered]
        \tikzstyle{connect}=[-latex, thick]
        
        \node[main,text width=3cm] (source) [] {$s_1, s_2, ..., s_{N_s}$};
        \node[main,text width=2cm] (spe) [above=of source] {Positional Encoding};
        \node[main,text width=2.5cm] (enc) [above=of spe] {Transformer Encoder};
          
        \node[main,text width=4cm, color=blue!100] (ptarget) [right=5 mm of source] {$e^{enc}_{z_1}, e^{enc}_{z_2}, ..., e^{enc}_{z_{N_Z}}$};
        
        \node[minimum height = 5.7cm, minimum width = 5.5cm, thick, draw =black!80, node distance = 5mm, font=\Large, align=left, dashed, yshift=1.4cm, color=blue!100] (dec) [above=of ptarget] {};
        \node[main,text width=5cm, align=left, yshift=1.6cm] (att) [above=of ptarget] {
        \textcolor{blue}{Cross-attention to align source information with latent variable information}};
        \node[main,text width=5cm] (mlp) [above=of att] {MLP};
        \node[minimum width = 2.8cm, thick, draw =black!80, node distance = 5mm, font=\Large, align=left, minimum height = 1cm,  text width=2.8cm, xshift=-12mm, dashed, color=blue!100] (declab) [above=of mlp] {Transformer\\ Encoder};
        
        \node[main,text width=2.8cm, xshift=-2.0cm, yshift=0.3cm, color=blue!100] (result1) [above=20 mm of mlp] {$\mu_1, \mu_2, ..., \mu_{N_Z}$};
        \node[main,text width=2.8cm, color=blue!100] (result2) [right=1mm of result1] {$\sigma_1, \sigma_2, ..., \sigma_{N_Z}$};
        
       \path (source) edge [connect] (spe)
            (spe) edge [connect] (enc)
            (enc) edge [bend left][connect] (att)
            (ptarget) edge [connect, color=blue!100] (att)
            (att) edge [connect] (mlp)
            (mlp) edge [bend right][connect, color=blue!100] (result1)
            (mlp) edge [bend right][connect, color=blue!100] (result2);
        \end{tikzpicture}}
    \end{adjustbox}
    \vskip -5px
    \caption*{(b) Our encoder}
    \end{minipage}\hfill
    \begin{minipage}[b]{0.33\linewidth}
    \begin{adjustbox}{minipage=\linewidth,scale=0.53}
        \fbox{\begin{tikzpicture}
        \tikzstyle{main}=[rounded corners, minimum size = 10mm, thick, draw =black!80, node distance = 5mm, font=\Large, text centered]
        \tikzstyle{connect}=[-latex, thick]
        
        \node[main,text width=3cm, color=blue!100] (source) [] {$z_1, z_2, ..., z_{N_Z}$};
        \node[main,text width=3cm, color=blue!100] (spe) [above=of source] {z-Identifier};
        \node[main,text width=2.5cm] (enc) [above=of spe] {Transformer Encoder};
          
        \node[main,text width=4cm] (ptarget) [right=5 mm of source] {$s_1, s_2, ..., s_{N_s-1}$};
        \node[main,text width=2cm] (tpe) [above=of ptarget] {Positional Encoding};
        
        \node[minimum height = 5.7cm, minimum width = 5.5cm, thick, draw =black!80, node distance = 5mm, font=\Large, align=left, dashed, yshift=-0.2cm] (dec) [above=of tpe] {};
        \node[main,text width=5cm, align=left] (att) [above=of tpe] {Masked Self Attention + Cross-attention to align latent variable information with source};
        \node[main,text width=5cm] (mlp) [above=of att] {MLP};
        \node[minimum width = 2.8cm, thick, draw =black!80, node distance = 5mm, font=\Large, align=left, minimum height = 1cm,  text width=2.8cm, xshift=-12mm, dashed] (declab) [above=of mlp] {Transformer\\ Decoder};
        
        \node[main,text width=5cm, yshift=0.1cm] (result) [above=20 mm of mlp] {$s_1, s_2, ..., s_{N_s}$};
        
       \path (source) edge [connect, color=blue!100] (spe)
            (spe) edge [connect] (enc)
            (enc) edge [bend left][connect] (att)
            (ptarget) edge [connect] (tpe)
            (tpe) edge [connect] (att)
            (att) edge [connect] (mlp)
            (mlp) edge [bend right][connect] (result);
        \end{tikzpicture}}
    \end{adjustbox}
    \vskip -5px
    \caption*{(c) Our decoder}
    \end{minipage}
    \caption{
    In blue, we highlight in (b) the difference between our encoder and a source-to-target MT model, and in (c) the difference between our decoder and a target-to-source MT model. The input at the bottom right for the Transformer Decoders in (a) and (c) is the series of previous words for autoregressive generation. The input to our model is a series of words $s$, at the bottom left of (b), and its output is the reconstruction of these words in the same language, at the top right of (c).
        }
    \label{fig:SCHEMES}
    
\end{figure*}

\subsection{The Intuition Behind our Model}
\label{MOTIVATION}
Consider $s=(s_j)_{1\leq j\leq N_s}$ and $t=(t_j)_{1\leq j\leq N_t}$, two series of tokens forming respectively a sentence in 
a source language and a sentence in a target language. Given $s$, attention-based translation models are capable of yielding 
$t$ while also providing information about the alignment between the groups of tokens (of different sizes) in both sentences 
\citep{Bahdanau2015NeuralTranslate,Luong2015EffectiveTranslation}). 
  This evidence suggests that attention-based architectures are capable of factoring information from groups  of words 
  according to a source structure, and redistributing it according to a target structure.

 The aim of our design is to use, as a target, a set  of $N_Z$ \emph{independent} latent variables that will act as fixed placeholders
  for the information in sentences. We stress that 
   $N_Z$ is fixed and independent of the input sentence size $N_s$. Combining Transformers, an attention-based MT model, 
 and the VAE framework for disentanglement,
 our ADVAE  is intended to factor information from independent groups of words into separate latent variables. 
 In the following sections, we will refer to this set of independent latent variables as the latent \textit{vector} $z=(z_i)_{1\leq i\leq N_Z}$ and to each $z_i$ as a latent \textit{variable}.
 
 \subsection{Model Architecture}
 \label{ARCHITECTURE}
\paragraph{Inference model:}
This is the inference model $q_\phi$ (encoder in Fig. \ref{fig:SCHEMES}.b) for our latent variables $z=(z_i)_{1\leq i\leq N_Z}$. It differs from an MT Transformer in two ways.
 First it uses as input a sentence $s$, and $N_Z$ learnable vectors $(e^{enc}_{z_i})_{1\leq i\leq N_Z}$ instead of the target tokens $t$ used in translation. These learnable vectors will go through Cross-Attention without Self-Attention. We stress that these learnable vectors are input-independent.
Second its output is not used to select a token from a vocabulary but rather
passed to a linear layer (resp. a linear layer followed by a softplus non-linearity) to yield the mean parameters $(\mu_i)_{1\leq i\leq N_Z}$ (resp.
 the standard deviation parameters  $(\sigma_i)_{1\leq i\leq N_Z}$)  to parameterize the diagonal Gaussian distributions $(q^{(i)}_\phi(z_i|s))_{1\leq i\leq N_Z}$. The Transformer Decoder is therefore replaced in Fig \ref{fig:SCHEMES}.b by a Transformer Encoder that uses Cross-attention to factor information
from the sentence.  The distribution of the whole latent vector is simply the product of Gaussians
 $q_\phi(z_1,\ldots,z_{N_{Z}} |s)=\prod_i^{N_Z}q^{(i)}_\phi(z_i|s)$. 

\paragraph{Generation model:}
Our generation model consists of an autoregressive decoder (Fig. \ref{fig:SCHEMES}.c) $p_\theta(s|z_1,\ldots, z_{N_{Z}}) = \prod_j^{N_s}p_\theta(s_j|s_{<j}, z_1,\ldots, z_{N_{Z}})$ where $s_{<i}$ is the series of tokens preceding $s_i$, and
a  prior assuming independent standard Gaussian variables, \emph{i.e.} $p(z_1,\ldots,z_{N_{Z}}) =  \prod_i^{N_Z}p(z_i)$. 

Each latent variable $z_i$ is concatenated with an associated learnable vector $e^{dec}_{z_i}$ (\emph{z-Identifier} in Fig. \ref{fig:SCHEMES}.c) instead of going through positional encoding. From there on, the latent variables are used like source tokens in an MT Transformer.

\subsection{Optimization Objective}
\label{OBJECTIVE}
We train our ADVAE using the 
$\beta$-VAE \citep{Higgins2019-VAE:Framework} objective, which is the Evidence Lower-Bound (ELBo) with a controllable weight 
on its Kullback-Leibler (KL) term:
\begin{equation}
    \log p_\theta(s) \geq \mathbb{E}_{(z) \sim q_\phi(z|s)}\left[ \log p_\theta(s|z) \right] -
    \beta \KL[q_\phi(z|s)||p(z)] \label{BETAVAE}
\end{equation}
In Eq. \ref{BETAVAE}, $s$ is a sample from our dataset, $z$ is our latent vector and the distributions 
$p_\theta(s) = \int p_\theta(s|z) p(z)dz$ and $q_\phi(z|s)$ are respectively the generation model and the inference model.
We use a standard Gaussian  distribution as prior $p(z)$ and a diagonal Gaussian distribution as the approximate
inference distribution $q_\phi(z|s)$. The weight $\beta$ is
used (as in \citealp{Chen2018c}, \citealp{Xu2020OnSupervision}, \citealp{Li2020ProgressiveRepresentations}) to control
 disentanglement, but also to find a balance between the expressiveness of latent variables 
 and the generation quality.

\section{Evaluation Protocol}
\label{EVALUATIONPROC}
In order to quantify disentanglement, we first measure the interaction between latent variables and syntactic roles. To do so, we extract \emph{core} syntactic roles from sentences according to the procedure we describe in \S \ref{ROLEEXTRACT}. Subsequently, for the ADVAE decoder, we repeatedly perturb latent variables and measure their influence on the realizations of the syntactic roles in generated sentences (\S\ref{DECODERMETRICS}). For the ADVAE encoder, we use attention to determine the syntactic role that participates most in  producing the value of each latent variable (\S\ref{ENCODERMETRICS}). 

Given these metrics, we measure disentanglement taking inspiration from the Mutual Information Gap (MIG; \citealp{Chen2018c}) in \S \ref{DISENTMETRICS}. MIG consists in measuring the difference between the first and second latent variables with the highest mutual information with regard to a target factor. It is intended to quantify the extent to which a target factor is concentrated in a single variable.
This metric assumes knowledge of the underlying distribution of the target information in the dataset.
However, there is no straightforward or agreed-upon way to set this distribution for text spans, and therefore to calculate MIG in our case. As a workaround, we use the influence metrics defined in \S\ref{DECODERMETRICS} and \S\ref{ENCODERMETRICS} as a replacement for mutual information to quantify disentanglement.

\subsection{Syntactic role extraction}\label{ROLEEXTRACT}
We use the Spacy\footnote{\url{https://spacy.io/models/en\#en\_core\_web\_sm}} dependency parser~\citep{spacy2}
trained on Ontonotes5~\citep{AB2/MKJJ2R_2013}. For each sentence the realization of \emph{verb} is
 the root of the dependency tree if its POS tag is \textit{VERB}. Realizations of \emph{subj}
(subject), \emph{dobj} (direct object), and \emph{pobj} (prepositional object) are \emph{spans}
of subtrees whose roots are labelled resp. \textit{nsubj}, \textit{dobj}, and \textit{pobj}.

In the rare cases where multiple spans answer the requirement for a syntactic role, we take the first one as the subsequent spans are most often part of a subordinate clause.
A realization of a syntactic role in $R=\{verb, subj, dobj, pobj\}$ is empty if no node in the dependency tree satisfies its
extraction condition.\footnote{Examples of syntactic role extractions can be found in Appendix~\ref{EXAMPLESEXTRACTIONS}.}

\subsection{Latent Variable Influence on Decoder}\label{DECODERMETRICS}

Intuitively, we repeatedly compare the text generated from a sampled latent vector to the text generated using the same \emph{vector} where only one latent \emph{variable} is resampled.
Thus we can isolate the effect of each latent $variable$ on output text and gather statistics.

More precisely, we sample $T^{dec}$ latent \emph{vectors} $(z^{(l)})_{1\leq l\leq T^{dec}}=(z_i^{(l)})_{1\leq l\leq T^{dec}, 1\leq i\leq N_Z}$.
Then for each $z^l$, and for each $i$ we create an altered version $\tilde{z}^{(li)}=(\tilde{z}^{(li)}_{i'})_{1\leq i'\leq N_Z}$ where we resample only the $i^{\text{th}}$ latent \emph{variable} (\textit{i.e.} $\forall i'\neq i,\  \tilde{z}^{(li)}_{i'}=z^{(l)}_{i'}$).

Generating the corresponding sentences\footnote{Throughout this work, we use greedy sampling (sampling the maximum-probability word at each step), for all generated sentences. } with $p_\theta(s|z)$ yields a list of original sentences ${(s^{(l)})}_{1\leq l\leq T^{dec}}$, and a matrix of sentences displaying the effect of modifying each latent variable ${(\tilde{s}^{(li)})}_{1\leq l\leq T^{dec}, 1\leq i\leq N_Z}$. 
For each syntactic role $r \in R$, we will denote the realization extracted from a sentence $s$ with $\rop(s)$.

To measure the influence of a variable $z_i$ on the realization of a syntactic role $r$, denoted $\Gamma^{dec}_{ri}$, we estimate the probability that a change in this latent variable incurs a change in the span corresponding to the syntactic role.
We first discard, for the influence on a role $r$, sentence pairs  $(s^{(l)}, \tilde{s}^{(li)})$ where it appears or disappears, 
because the presence of a syntactic role is a property of its parent word, 
(\textit{e.g.} the presence or absence  of  a \emph{dobj} is controlled by the \textit{transitivity} of the verb) hence not directly connected to the representation of the role $r$ itself.
As they are out of the scope of our work, we report measures of these structural changes (diathesis) in Appendix \ref{StructHeatMap}, and leave their extensive study to future works. We denote the remaining number of samples $T'^{dec}_{ri}$.

In the following, we use operator $\mathbf{1}{\{.\}}$, which is equal to 1 when the boolean expression it contains is true and to 0 when it is false.
This process yields a matrix $\Gammaopdec$ of shape $(|R|, N_Z)$ which summarizes interactions in the \emph{decoder} between syntactic roles and latent variables: 
\begin{align}
    &\Gamma^{dec}_{ri} =
    \sum_{l=1}^{T'^{dec}_{ri}} \frac{ \mathbf{1}{\{\displaystyle\rop(s^{(l)})\neq \rop(\tilde{s}^{(li)})\}}}{T'^{dec}_{ri}} 
\end{align}
\subsection{Encoder Influence on Latent Variables}\label{ENCODERMETRICS}
We compute this on a held out set of size $T^{enc}$ of sentences ${(s^{(l)}_{j})}_{1\leq l\leq T^{enc}, 1\leq j\leq N_{s^{(l)}}}$.
Each sentence $s^{(l)}$ of size $N_{s^{(l)}}$ generates an attention matrix ${(a^{(l)}_{ij})}_{1\leq i\leq N_Z, 1\leq j\leq N_{s^{(l)}}}$. 
Attention values are available in the Transformer Encoder with cross-attention computing the inference model\footnote{For simplicity, attention values are averaged over attention heads and transformer layers. This also allows drawing conclusions with regard to the tendency of the whole attention network, and not just particular specialized heads as was done in \citet{Clark2019WhatAttentionb}. Nevertheless, we display per-layer results in Appendix \ref{PerLayerAtt}.}, and quantify the degree to which each latent variable embedding $e_{z_i}^{enc}$ draws information from each token $s_j$ to form the value of $z_i$.

For the encoder, we consider the influence of a syntactic role on a latent variable to be the probability for the attention values
of the latent variable to reach their maximum on the index of a token in that syntactic role's realization. The indices of tokens belonging to a syntactic role $r$ in a sentence $s^{(l)}$ are denoted $\argr(s^{(l)})$. For each syntactic role $r$ and sentence $s^{(l)}$, we discard 
inputs where this syntactic role cannot be found, and denote the remaining number of samples $T'^{enc}_{r}$. The resulting measure of influence of syntactic role $r$ on variable $z_i$ is denoted $\Gamma^{enc}_{ri}$.
The whole process yields matrix $\Gammaopenc$ of shape $(|R|, N_Z)$ which summarizes interactions in the \emph{encoder} between syntactic roles and latent variables:
\begin{align}
    &\Gamma^{enc}_{ri} =
    \sum_{l=1}^{T'^{enc}_{r}} \frac{\mathbf{1}{\{\displaystyle \argmax_{j}(a^{(l)}_{ij})\in \argr(s^{(l)})\}}}{T'^{enc}_{ r}}
\end{align}

\subsection{Disentanglement Metrics}\label{DISENTMETRICS}

For $\Gamma^*$ (either $\Gammaopdec$ or $\Gammaopenc$) each line corresponds to a syntactic role in the data.
The disentanglement metric for role $r$ is the following:
\begin{align}
    \Delta\Gamma^*_{r} &= \Gamma^*_{rm_1} -\Gamma^*_{rm_2}\label{DELTA}\\
    s.t. \hspace{0.2cm}
    m_1 &= \argmax_{1\leq i\leq N_Z} \Gamma^*_{ri}, \hspace{0.2cm}
    m_2 =  \argmax_{1\leq i\leq N_Z, j\neq m1} \Gamma^*_{ri}\nonumber
\end{align}

We calculate total disentanglement scores for syntactic roles using  $\Gammaopdec$, $\Gammaopenc$ as follows:
\begin{align}
    \mathbb{D}_{dec} = \sum_{r \in R} \Delta{\Gamma}^{enc}_{r}\hskip 1mm, \hskip 2mm 
    \mathbb{D}_{enc} = \sum_{r \in R} \Delta{\Gamma}^{enc}_{r} 
\end{align}

In summary, the more each syntactic role's information is concentrated in a single variable, the higher the values of $\mathbb{D}_{dec}$ and $\mathbb{D}_{enc}$.
However, similar to MIG, these metrics do not say whether variables capturing our concepts of interest are \emph{distinct}. Therefore, we also report the number of distinct variables that capture the most
each syntactic role (\textit{i.e} the number of distinct values of $m_1$ in
Eq.~\ref{DELTA} when looping over $r$). This is referred to as $N_{\Gammaopenc}$
for the encoder and $N_{\Gammaopdec}$ for the decoder.

\section{Experiments}
\label{EXPE}
\paragraph{Dataset} Previous unsupervised disentanglement works 
\citep{Higgins2019-VAE:Framework, Kim2018DisentanglingFactorising, Li2020ProgressiveRepresentations} tend to use relatively homogeneous 
and low complexity data.
The data has \textit{low complexity} if it varies along clear factors which correspond to what the model aims 
to disentangle.
Similarly, we use a dataset where samples exhibit low variance in terms of syntactic structure while providing a high diversity of realizations for the syntactic roles composing the sentences, which is an adequate test-bed for unsupervised disentanglement of syntactic roles' realizations. This dataset is the plain text from the SNLI dataset \citep{bowman-etal-2015-large}  extracted\footnote{
\url{github.com/schmiflo/crf-generation/blob/master/generated-text/train}}  by \citet{Schmidt2020AutoregressiveLoops}.
The SNLI data is a collection of premises (on average $8.92\pm 2.66$ tokens long) made for Natural Language Inference. We use 90K samples as a training set, 5K for development, and 5K as a test set.

\paragraph{Setup}
Our objective is to check whether the architecture of our ADVAE induces better syntactic role disentanglement. We compare it to standard Sequence VAEs 
\citep{Bowman2016GeneratingSpace} and to a Transformer-based baseline that doesn't use cross-attention. Instead of cross-attention, this second baseline uses mean-pooling over the output of a Transformer encoder for encoding. For decoding, it uses the latent variable as a first token in a Transformer decoder, as is done for conditional generation with GPT-2 \cite{santhanam-shaikh-2019-emotional}.
These comparisons are performed using the same $\beta$-VAE objectives, and the decoder disentanglement scores as metrics.  Training specifics and hyper-parameter settings are 
 detailed in Appendix \ref{TRAINING&HP}.  For each of the two baselines, the latent variables we vary during the decoder's 
 evaluation are the mono-dimensional components of its latent vector. It is easier to pack information about the realizations of multiple syntactic roles into $D_z$ dimensions than into a single dimension. Consequently, the single dimensions we study for the baselines should be at an advantage to separate information into different variables.

Scoring disentanglement on the encoder side will not be possible for the baselines above as it requires attention values. 
To establish that our model effectively tracks syntactic roles, we compare it to a third baseline that locates each syntactic role 
through its median position across the dataset. This baseline is fairly strong on a language where word order is rigid (\textit{i.e} configurational language) such as English. We refer to this Position Baseline as PB.

The scores are given for different values of $\beta$ (Eq. \ref{BETAVAE}). Raising $\beta$ lowers the expressiveness of
 latent variables, but yields better disentanglement \citep{Higgins2019-VAE:Framework}. 
Following \citet{Xu2020OnSupervision}, we set $\beta$ to low values to avoid posterior collapse. In our case,
we observed that the models do not collapse for $\beta<0.5$. Therefore, we display results for $\beta \in \{0.3, 0.4\}$. We stop at 0.3 as lower values for $\beta$ result in poorer generation quality.
For our model we report performance for instances with $N_Z=4$ (\emph{ours-4}) and $N_Z=8$ (\emph{ours-8}).

\paragraph{Results}
The global disentanglement metrics are reported in Table \ref{tab:results}.\footnote{Fine-grained scores are given in Appendix~\ref{FULLSYNRESULTS}.}
  \begin{table}[!htbp]
    \centering
    \caption{Disentanglement quantitative results for the encoder (enc) and the decoder (dec). $N_{\Gamma}$ indicates the number of separated syntactic roles, and $\mathbb{D}$ measures concentration in a single variable. Values are averaged over 5 experiments. The standard deviation is between parentheses.}
    \resizebox{0.6\textwidth}{!}{%
    \begin{tabular}{|c|c||c|c||c|c|}
    \hline
    Model& $\beta$ & $\mathbb{D}_{enc}\uparrow$ &  $N_{\Gammaopenc}\uparrow$  &  $\mathbb{D}_{dec}\uparrow$ &  $N_{\Gammaopdec}\uparrow$\\
    \hline \hline
    
    \multirow{2}{*}{Sequence VAE}&0.3& -& -& 0.60\textcolor{gray}{(0.09)}& 2.40\textcolor{gray}{(0.55)}\\ 
    &0.4& -& -& 1.28\textcolor{gray}{(0.24)}& 1.40\textcolor{gray}{(0.55)}\\
    \hline
    \multirow{2}{*}{Transformer VAE}&0.3& -& -& 0.12\textcolor{gray}{(0.10)}& 3.00\textcolor{gray}{(0.70)}\\ 
    &0.4& -& -& 0.11\textcolor{gray}{(0.04)}& 3.20\textcolor{gray}{(0.44)}\\
    \hline
    \multirow{1}{*}{PB}& - & 0.98 \textcolor{gray}{(-)}& 3.00\textcolor{gray}{(-)}& - & - \\     \hline
    \hline
    \multirow{2}{*}{ours-4} &0.3& 1.48\textcolor{gray}{(0.15)}& 3.00\textcolor{gray}{(0.00)}& 0.71\textcolor{gray}{(0.06)}& 3.00\textcolor{gray}{(0.00)}\\ 
     &0.4& 1.43\textcolor{gray}{(0.79)}& 3.00\textcolor{gray}{(0.00)}& 0.72\textcolor{gray}{(0.37)}& 2.80\textcolor{gray}{(0.45)}\\
    \hline
    \multirow{2}{*}{ours-8} &0.3& 1.34\textcolor{gray}{(0.18)}& 3.80\textcolor{gray}{(0.45)}& 0.51\textcolor{gray}{(0.14)}& 2.80\textcolor{gray}{(0.45)}\\ 
     &0.4& 1.75\textcolor{gray}{(0.47)}& 2.80\textcolor{gray}{(0.45)}& 0.98\textcolor{gray}{(0.27)}& 2.60\textcolor{gray}{(0.89)}\\

 \hline
     \end{tabular}}
    \label{tab:results}
  \end{table}
  
\begin{figure*}[!h]
\centering
    \begin{minipage}[b]{0.45\textwidth}
            \centering
           \begin{minipage}[b]{\textwidth}
            \begin{adjustbox}{minipage=\textwidth,scale=0.45}
            \hspace{ 1cm} \includegraphics[trim={1.3cm 0.7cm 2.2cm 1.3cm},clip] {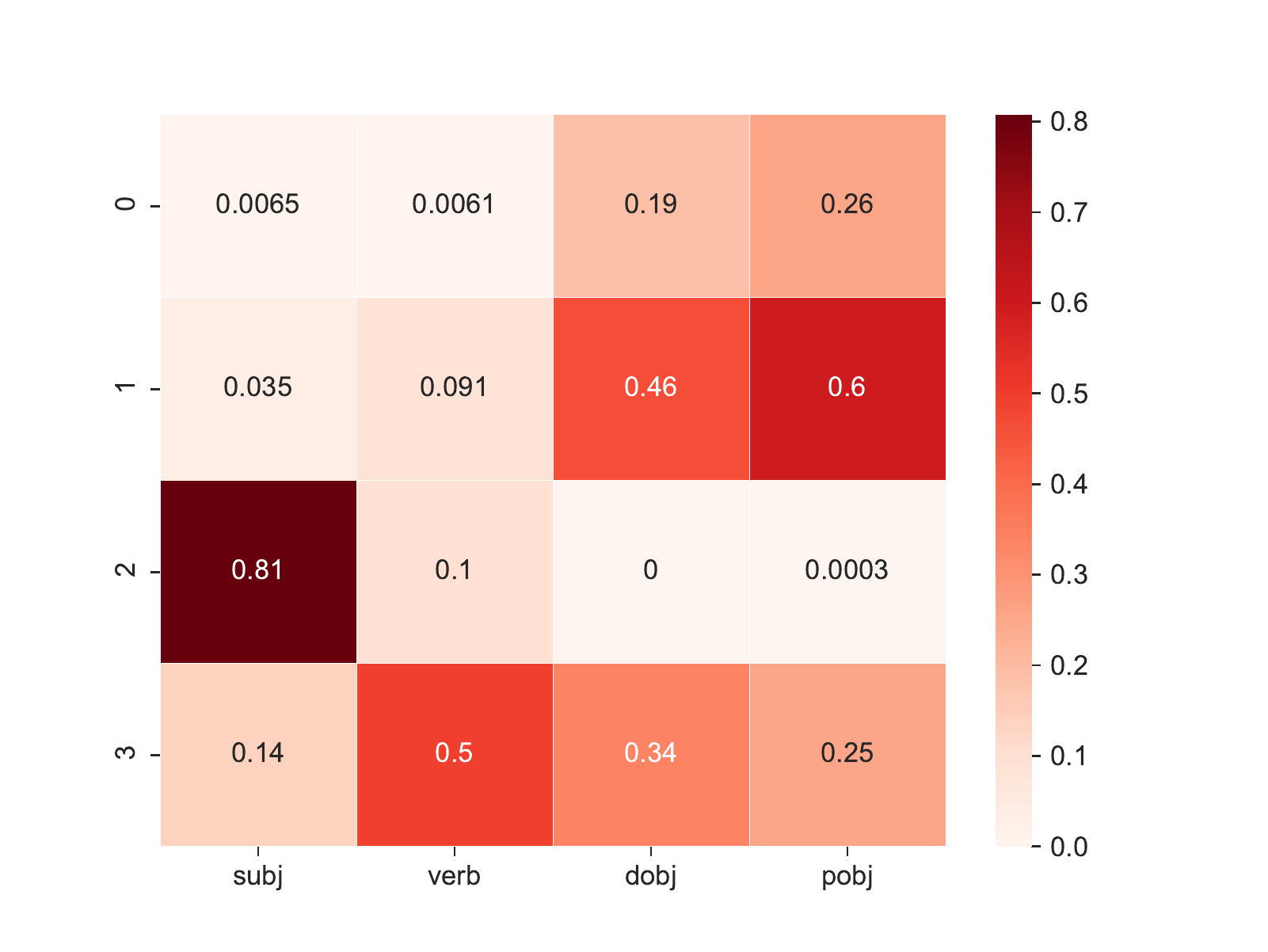}
            \end{adjustbox}
            \end{minipage}
            \caption{\centering Encoder influence heatmap ($\Gammaopenc$).}
            \label{fig:ENCHEAT}
    \end{minipage}
    \begin{minipage}[b]{0.45\textwidth}
            \centering
            \begin{minipage}[b]{\textwidth}
            \begin{adjustbox}{minipage=\textwidth,scale=0.45}
             \hspace{ 1cm} \includegraphics[trim={1.4cm 0.7cm 2.2cm 1.3cm},clip] {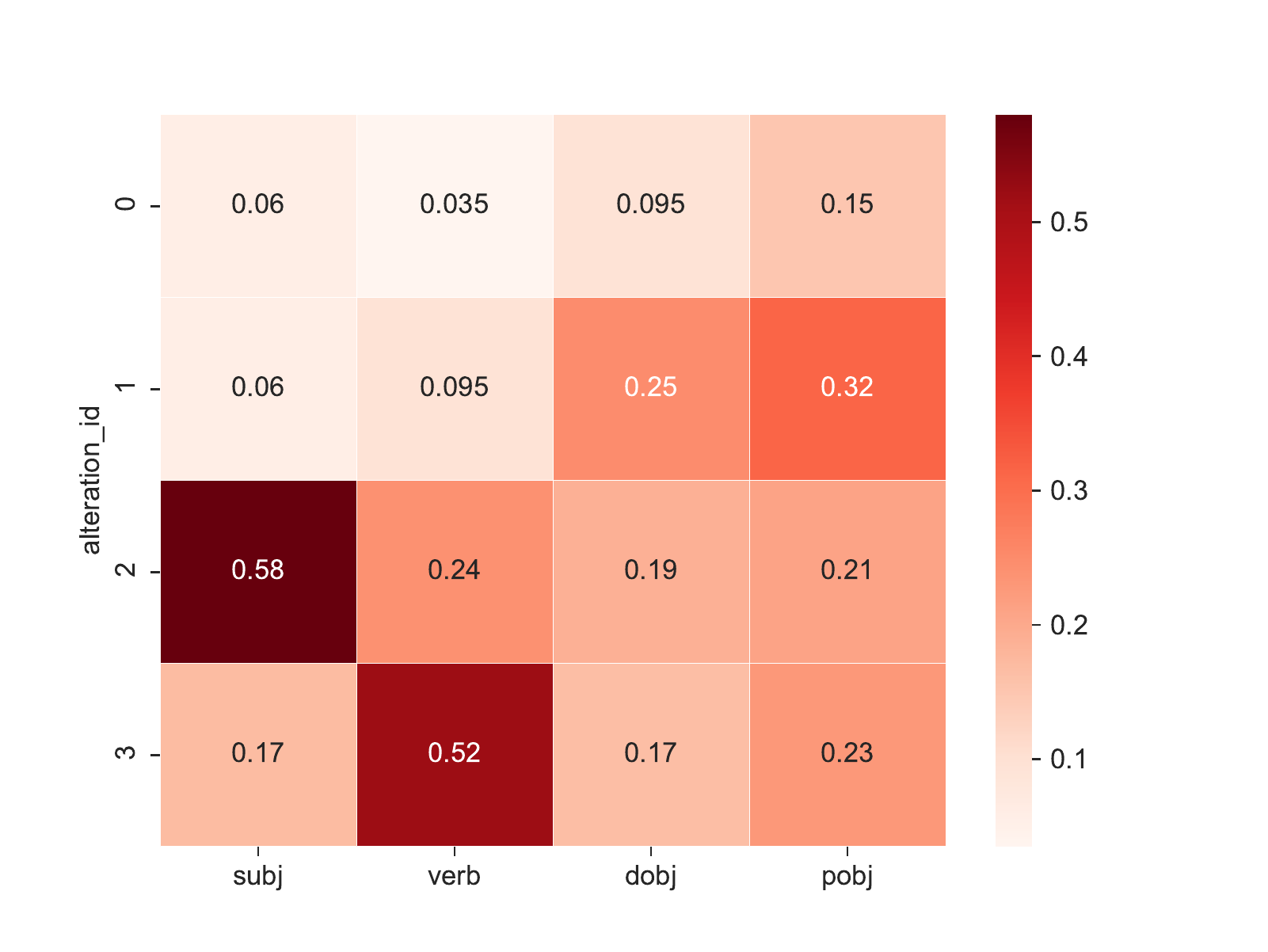}
            \end{adjustbox}
            \end{minipage}
            \caption{\centering Decoder influence heatmap ($\Gammaopdec$).}
            \label{fig:DECHEAT}
    \end{minipage}
\end{figure*} 
On the decoder side, the Sequence VAE exhibits disentanglement scores in the range of those reported for our model for $\beta=0.3$, and higher for $\beta=0.4$.
However, $N_{\Gammaopdec}$ shows that it struggles to factor the realizations of different syntactic roles in different latent variables,
 and the higher score shown for $\beta=0.4$ is accompanied by a lower tendency to separate the information from different syntactic 
 roles. The Transformer VAE baseline assigns different latent variables to the different syntactic role (high $N_{\Gammaopdec}$), but suffers from very low specialization for these latent variables (low $\mathbb{D}_{dec}$). 
In contrast, our model is consistently able to separate 3 out of 4 syntactic roles, and while a higher $\beta$ raises its $\mathbb{D}_{dec}$, it does not decrease its $N_{\Gammaopdec}$.
As \emph{ours-8} has more latent variables, this encourages the model to further split the information in each syntactic role between more latent variables\footnote{Results for a larger grid of $N_z$ values are reported in Appendix \ref{NZVARY}, and show that latent variables still clearly relate to syntactic roles, but in groups.}.
The fact that ADVAEs perform better than both Sequence VAEs and classical Transformer VAEs shows that  its disentanglement capabilities are due to the usage of Cross-Attention to obtain latent variables, and not only to the usage of Transformers. On the encoding side, our models consistently score above the baseline, showing that our latent variables
 actively follow the syntactic roles.

In Figures \ref{fig:ENCHEAT} and \ref{fig:DECHEAT}, we display the influence matrices $\Gammaopenc$ and $\Gammaopdec$ for an 
instance of our ADVAE with $N_Z=4$ as heatmaps.
The vertical axes correspond to the latent variables.
As can be seen, our model successfully associates latent variables to verbs and subjects 
but chooses not to separate direct objects and prepositional objects into different latent variables.
Upon further inspection of the same heatmaps for the VAE baseline, it appears that it most often uses a single latent variable for 
\emph{verb} and \emph{subj}, and another for \emph{dobj} and \emph{pobj}.

One can also notice in Figures \ref{fig:ENCHEAT} and \ref{fig:DECHEAT}, that the encoder matrix is sparser than the decoder matrix 
(which is consistent with the higher encoder disentanglement scores in Table \ref{tab:results}).
This is to be expected as the decoder $p_\theta(s|z)$ adapts the realizations of syntactic roles to each other after they are
 sampled separately from $p(z)$.
The reason for this is that the language modeling objective requires some coherence between syntactic roles
 (conjugating verbs with subjects, changing objects that are semantically inadequate for a verb, etc).
This \emph{co-adaptation}, contradicts the independence of our latent variables.
It will be further discussed in the following paragraph.

\begin{table*}[t]
    \small
    \centering
    \caption{Resampling a specific latent variable for a sentence. The ID column is an identifier for the example.}
    \begin{tabularx}{14cm}{|c|X|X|X|X|}
    \hline
     ID&Original sentence& Resampled subject& Resampled verb & Resampled dobj/pobj \\
    \hline \hline
     1& people are sitting on the beach & a young man is sitting on the beach & people are playing in the beach  & people are sitting on a bench
    \\\hline
     2&a man and woman are sitting on a couch & a man is sitting on a park bench &a man and woman are running on a grassy field   & the man and woman are on a beach\\\hline
     3&a man is playing with his dog & a boy is playing in the snow & a man is selling vegetables & a man is playing the game with his goal .\\\hline
     \end{tabularx}
    \label{tab:resultsresample}
\end{table*} 
\begin{table*}[t]
    \small
    \centering
    \caption{Swapping the value of a specific latent variable between two sentences. The SSR (Swapped Syntactic Role) column indicates the syntactic role that has been swapped.}
    \begin{tabularx}{14cm}{|p{0.2cm}|X|X|p{0.8cm}|X|X|}
    \hline
     ID& Sentence 1& Sentence 2& SSR & Swapped Sentence 1 & Swapped Sentence 2 \\
    \hline \hline
      1& a woman is talking on a train & people are sitting on the beach & subj & people are talking on a train & a woman is sitting on the beach \\\hline
      2&people are sitting on the beach  &   a woman is talking on a train& verb& people are talking on a beach&   a woman is standing on a train \\\hline
      3&a woman is talking on a train &   a man is playing with his dog  & \makecell{dobj/\\pobj}  & a man is playing the guitar with a goal & a woman is performing a trick\\\hline
      \end{tabularx}
    \label{tab:resultsswap}
\end{table*} 
\paragraph{Changing the Realizations of Syntactic Roles}
\label{QUALITATIVE}
Here, we display of few qualitative examples of how the realizations of syntactic roles can be separately changed using an instance of our ADVAE.

As a first example, we generate a sentence from a random latent vector, then resample for each syntactic role the corresponding
 disentangled latent variable to observe the change on the subsequently generated altered sentence. The results of this manipulation 
 are in Table~\ref{tab:resultsresample}\footnote{More Examples are available in Appendix \ref{QUALIAPPEN}}. As can be seen, some examples exhibit changes that only affect the target syntactic role 
 (example 1).
However, the model often produces co-adaptations that go past the target syntactic role either for semantic soundness
 (example 2, resampled verb adapts the object), or simply for lack of generalization 
 from 
 the SNLI data used for training.

A second example we display is a swap of syntactic role  realizations between sentences.
A few examples are given in Table~\ref{tab:resultsswap}. Similar to Table~\ref{tab:resultsresample}, the model often
yields the expected result. Co-adaptation is best seen here, as taking a syntactic role to a sentence with which 
it is incompatible results in unexpected changes (example 3).

\paragraph{Further investigations}
As this is a first step in this research direction, we conducted this study on a dataset of relatively regular sentences. Running similar experiments on a dataset with more complicated and diverse sentence structures such as in Yelp (Appendix \ref{YELP}) results in the same comparative patterns. However, disentanglement scores are much lower. This calls for future iterations to improve upon ADVAE and our evaluation protocol to better model structure in order to scale to User Generated Content (UGC). Our experiments also enabled underlining an inherent issue to syntactic role disentanglement: \emph{co-adaptation}. The independence between our latent variables causes the decoder $p_\theta(s|z)$ to correct the incoherence between independently sampled syntactic role realizations. Using structured latent variables to learn relations between syntactic roles seems to be the natural solution to this problem. An investigation of a hierarchical version of the ADVAE (Appendix \ref{HIERARCH}) showed, however, that a drop-in replacement of the independent prior with a structured prior is not sufficient in order to \emph{absorb} co-adaptation into the latent variable model. Our future works will, therefore, also include the investigation of training techniques that can achieve improved results with structured latent variables.
   
\section{Conclusion}
We introduce a framework to study the disentanglement of  syntactic roles and show that it is possible to learn a representation of sentences that exhibits separation in the realizations of these syntactic functions \emph{without supervision}. Our framework includes: \textit{i)} Our model, the ADVAE, which maps syntactic roles to separate latent variables more often than standard Sequence VAEs and with better concentration than standard Transformer VAEs, and allows for the use of attention to study the interaction between latent variables and spans, \textit{ii)} An evaluation protocol to quantify disentanglement between latent variables and spans both in the encoder and in the decoder.

Our study constitutes a first step in a promising process towards \emph{unsupervised} explainable modeling and fine-grained control over the predicate-argument structure of sentences. Although we focused on syntactic roles realizations, this architecture as well as the evaluation method are generic and could be applied to other tasks. The architecture could be used at the document level (\emph{e.g.} disentangling discourse relations), while the evaluation protocol could be applied to other spans such as constituents.

 \subsubsection*{Acknowledgments}
This work is supported by the PARSITI project grant (ANR-16-CE33-0021) given by the French National Research Agency (ANR), the \emph{Laboratoire d’excellence “Empirical Foundations of Linguistics”} (ANR-10-LABX-0083), as well as the ONTORULE project. It was also granted access to the HPC resources of IDRIS under the allocation 20XX-AD011012112 made by GENCI.


\bibliography{references, bib2}
\bibliographystyle{iclr2022_conference}
\clearpage
\appendix

\section{A Hierarchical Version of our ADVAE}
\label{HIERARCH}
As we stated, our ADVAE aims to factor sentences into independent latent variables.
However, given the dependency structure of sentences,  realizations of syntactic roles are known to be interdependent to some degree in general. 
Therefore one may think that a structured latent variable model would be better suited to model the realizations of syntactic roles. In fact, such a model could absorb the language modeling co-adaptation between syntactic roles. For instance, instead of sampling an object and a verb from $p(z)$ that are inadequate, then co-adapting them through $p_\theta(s|z)$, a structured $p_\theta(z)$ could  produce an \emph{adequate} object for the verb.
For this experiment, rather than using an independent prior $p(z)$, we use a structured prior 
$p_\theta(z) = p(z^0) \prod^L_{l=1} p_\theta(z^l|z^{l-1})$  where $p(z^0)$ is a standard Gaussian, and all subsequent
 $L-1$ hierarchy levels are parameterized by learned conditional diagonal Gaussians. The model used for each $p_\theta(z^l|z^{l-1})$ 
 is shown in Figure \ref{fig:DEEPSCHEME} below:
\begin{figure}[!h]
\centering
    \begin{minipage}[b]{0.33\textwidth}
    \begin{adjustbox}{minipage=\textwidth,scale=0.6}
        \fbox{\begin{tikzpicture}
        \tikzstyle{main}=[rounded corners, minimum size = 10mm, thick, draw =black!80, node distance = 5mm, font=\Large, text centered]
        \tikzstyle{connect}=[-latex, thick]
        
        \node[main,text width=3cm, color=blue!100] (source) [] {$z^{l-1}_{1}, z^{l-1}_{2},$ \\$..., z^{l-1}_{N_Z}$};
        \node[main,text width=2cm, color=blue!100] (spe) [above=of source] {z-identifier};
        \node[main,text width=2.5cm] (enc) [above=of spe] {Transformer Encoder};
          
        \node[main,text width=4cm, color=blue!100] (ptarget) [right=5 mm of source] {$e_{z^{l}_{1}}^{enc}, e_{z^{l}_{2}}^{enc},..., e_{z^{l}_{N_Z}}^{enc}$};
        
        \node[minimum height = 5.7cm, minimum width = 5.5cm, thick, draw =black!80, node distance = 5mm, font=\Large, align=left, dashed, yshift=1.4cm] (dec) [above=of ptarget] {};
        \node[main,text width=5cm, align=left, yshift=1.6cm] (att) [above=of ptarget] {\textcolor{blue}{Self Attention} + Cross-attention to align source information with latent variables};
        \node[main,text width=5cm] (mlp) [above=of att] {MLP};
        \node[minimum width = 2.8cm, thick, draw =black!80, node distance = 5mm, font=\Large, align=left, minimum height = 1cm,  text width=2.8cm, xshift=-12mm, dashed] (declab) [above=of mlp] {Transformer\\ Decoder};
        
        \node[main,text width=3.4cm, xshift=-2.3cm, yshift=0.1cm, color=blue!100] (result1) [above=20 mm of mlp] {$\mu^{l}_{1}, \mu^{l}_{2}, ..., \mu^{l}_{N_Z}$};
        \node[main,text width=3.4cm, color=blue!100] (result2) [right=1mm of result1] {$\sigma^{l}_{1}, \sigma^{l}_{2}, ..., \sigma^{l}_{N_Z}$};
        
       \path (source) edge [connect, color=blue!100] (spe)
            (spe) edge [connect] (enc)
            (enc) edge [bend left][connect] (att)
            (ptarget) edge [connect, color=blue!100] (att)
            (att) edge [connect] (mlp)
            (mlp) edge [bend right][connect, color=blue!100] (result1)
            (mlp) edge [bend right][connect, color=blue!100] (result2);
        \end{tikzpicture}}
    \end{adjustbox}
    \vskip -5px
    \caption*{(b)}
    \end{minipage}
    \caption{The conditional inference module linking each of the hierarchy levels in our prior with the next level $p_\theta(z^l|z^{l-1})$. This module treats latent variables from previous layers as they are treated in our original decoder, and generates parameters for latent variables in subsequent hierarchy levels as it is done in our encoder.}
    \label{fig:DEEPSCHEME}
    
\end{figure}

We display the results for $L=2$ and $L=3$ in Table \ref{tab:resultsDeep}. For both models, we set $N_Z$ to 4.

\begin{table}[!h]
    \centering
    \caption{Disentanglement results for structured latent variable models on SNLI. }
    \resizebox{0.49\textwidth}{!}{%
    \begin{tabular}{|c|c||c|c||c|c|}
    \hline
    Depth & $\beta$ & $\mathbb{D}_{enc}$ &  $N_{\Gammaopenc}$  &  $\mathbb{D}_{dec}$ &  $N_{\Gammaopdec}$\\
    \hline \hline
    \multirow{2}{*}{$L=2$} &0.3& 0.79\textcolor{gray}{(0.36)}& 3.60\textcolor{gray}{(0.55)}& 0.51\textcolor{gray}{(0.22)}& 2.60\textcolor{gray}{(0.55)}\\
     &0.4& 0.42\textcolor{gray}{(0.23)}& 2.80\textcolor{gray}{(0.45)}& 0.12\textcolor{gray}{(0.20)}& 2.20\textcolor{gray}{(0.45)}\\
    \hline
    \multirow{2}{*}{$L=3$} &0.3& 0.90\textcolor{gray}{(0.25)}& 3.14\textcolor{gray}{(0.69)}& 0.52\textcolor{gray}{(0.20)}& 2.43\textcolor{gray}{(0.53)}\\
     &0.4& 0.32\textcolor{gray}{(0.38)}& 2.75\textcolor{gray}{(0.50)}& 0.25\textcolor{gray}{(0.42)}& 2.25\textcolor{gray}{(0.50)}\\

 \hline
     \end{tabular}}
    \label{tab:resultsDeep}
  \end{table}
The results show lower mean disentanglement scores, and high standard deviations compared to the standard version of our ADVAE.
By inspecting individual training instances of this hierarchical model, we found that some instances achieve disentanglement
with close scores to those of the standard ADVAE, while others completely fail (which results in the high variances observed 
in Table \ref{tab:resultsDeep}). 
Unfortunately, hierarchical latent variable models are notoriously difficult to train \citep{Zhao2017LearningModels}. Our independent latent variable model is therefore preferable to the structured one due to these empirical results. More advanced hierarchical latent variable training techniques (such as Progressive Learning and Disentanglement  ~\citep{Li2020ProgressiveRepresentations}) may, however, provide better results.

\section{Experimenting with the Yelp Dataset}
\label{YELP}
 
\begin{figure*}[!h]
\centering
    \begin{minipage}[b]{0.48\textwidth}
            \centering
           \begin{minipage}[b]{\textwidth}
            \begin{adjustbox}{minipage=\textwidth,scale=0.5}
            \hspace{ 1cm} \includegraphics[trim={1.3cm 0.7cm 2.2cm 1.3cm},clip] {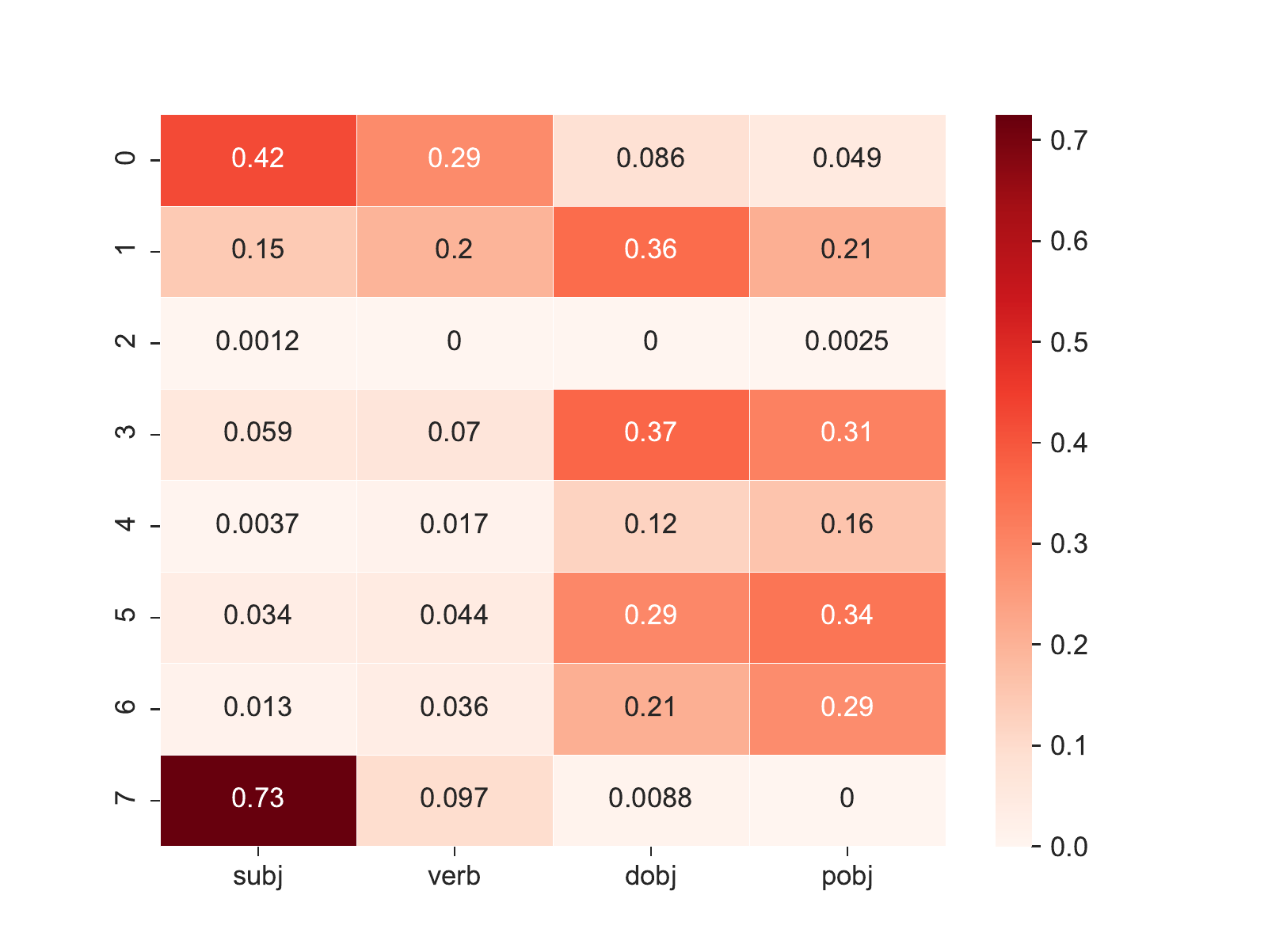}
            \end{adjustbox}
            \end{minipage}
            \caption{\centering Encoder influence heatmap for Yelp($\Gammaopenc$).}
            \label{fig:ENCHEATYELP}
    \end{minipage}
    \begin{minipage}[b]{0.48\textwidth}
            \centering
            \begin{minipage}[b]{\textwidth}
            \begin{adjustbox}{minipage=\textwidth,scale=0.5}
             \hspace{ 1cm} \includegraphics[trim={1.3cm 0.7cm 2.2cm 1.3cm},clip] {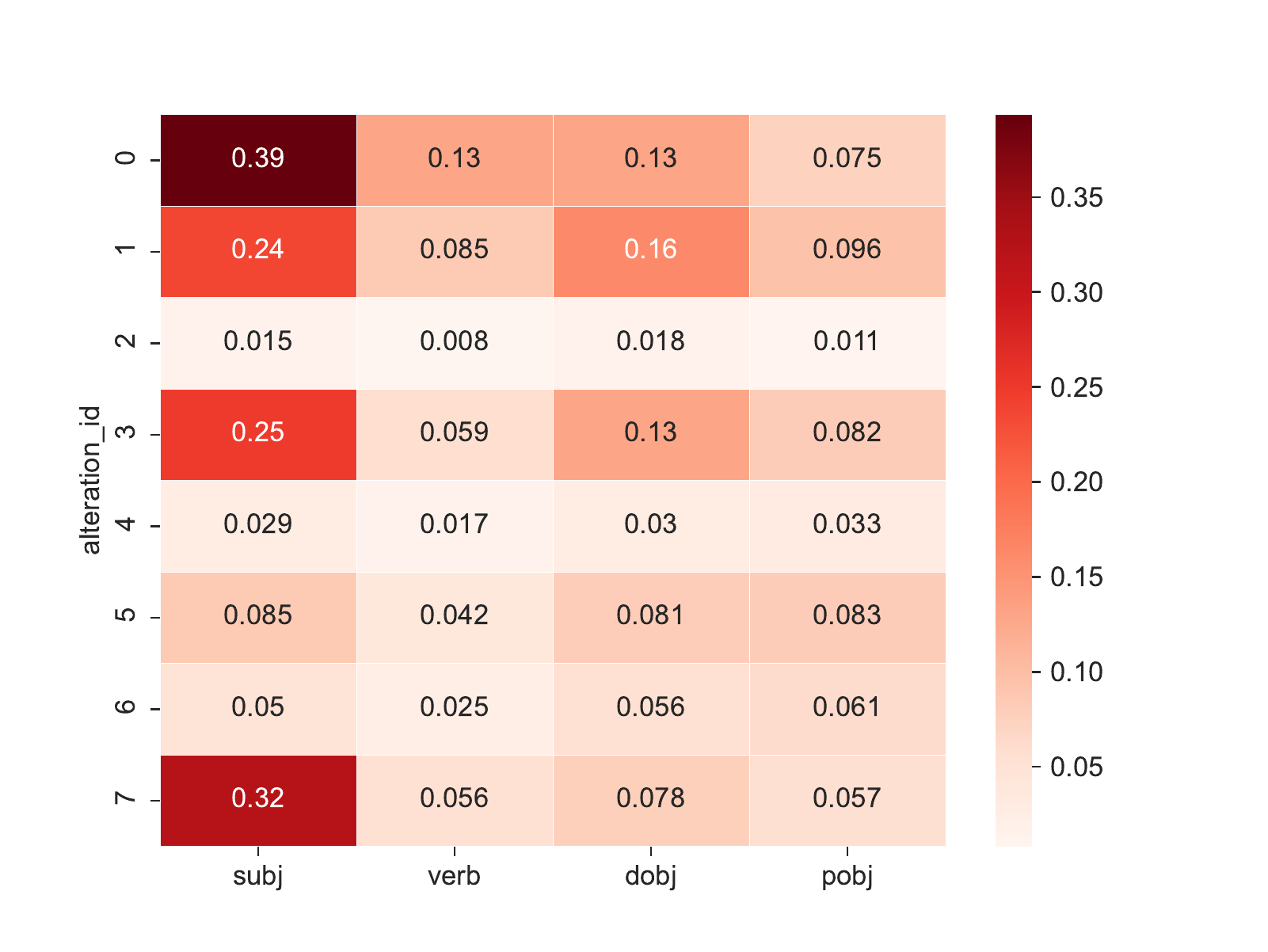}
            \end{adjustbox}
            \end{minipage}
            \caption{\centering Decoder influence heatmap for Yelp($\Gammaopdec$).}
            \label{fig:DECHEATYELP}
    \end{minipage}
\end{figure*}

We investigated the behavior of our ADVAE of on the user-generated reviews from the Yelp dataset used in \citet{li-etal-2018-delete}
 using the same procedure we used for SNLI. The length of sentences from this dataset ($8.88\pm3.64$) is similar to the length of sentences from the SNLI dataset.
Similar to the experiments in the main body of the paper, we display the disentanglement scores in Table \ref{tab:resultsYelp}, 
and the influence metrics of one of the instances of our model as heatmaps in Figures \ref{fig:ENCHEATYELP} and \ref{fig:DECHEATYELP}.

\begin{table}[!h]
    \centering
    \caption{Disentanglement results for the Yelp dataset}
    \resizebox{0.49\textwidth}{!}{%
    \begin{tabular}{|c|c||c|c||c|c|}
    \hline
    Model& $\beta$ & $\mathbb{D}_{enc}$ &  $N_{\Gammaopenc}$  &  $\mathbb{D}_{dec}$ &  $N_{\Gammaopdec}$\\
    \hline \hline
    \multirow{2}{*}{Sequence VAE}& 0.3 & - & - & 0.44\textcolor{gray}{(0.09)}& 2.20\textcolor{gray}{(0.45)}\\
    & 0.4&- & - & 1.21\textcolor{gray}{(0.06)}& 2.25\textcolor{gray}{(0.50)}\\
    \hline
    \multirow{2}{*}{Transformer VAE}&0.3& -& -& 0.17\textcolor{gray}{(0.12)}& 2.20\textcolor{gray}{(0.45)}\\ 
    &0.4& -& -& 0.23\textcolor{gray}{(0.03)}& 1.6\textcolor{gray}{(0.55)}\\
    \hline
    \multirow{1}{*}{PB}& - & 0.33\textcolor{gray}{(-)} & 2.00\textcolor{gray}{(-)} & - & - \\
    \hline \hline
    \multirow{2}{*}{ours-4} &0.3& 0.48\textcolor{gray}{(0.07)}& 2.00\textcolor{gray}{(0.00)}& 0.18\textcolor{gray}{(0.02)}& 2.50\textcolor{gray}{(0.58)}\\
     &0.4& 0.54\textcolor{gray}{(0.04)}& 3.00\textcolor{gray}{(0.00)}& 0.23\textcolor{gray}{(0.03)}& 2.40\textcolor{gray}{(0.55)}\\
    \hline
    \multirow{2}{*}{ours-8} &0.3& 0.44\textcolor{gray}{(0.04)}& 3.80\textcolor{gray}{(0.45)}& 0.17\textcolor{gray}{(0.04)}& 2.80\textcolor{gray}{(0.84)}\\
     &0.4& 0.57\textcolor{gray}{(0.26)}& 3.40\textcolor{gray}{(0.55)}& 0.15\textcolor{gray}{(0.10)}& 2.40\textcolor{gray}{(0.89)}\\

 \hline
     \end{tabular}}
    \label{tab:resultsYelp}
  \end{table}

Although the results show similar trends, they are weaker than what we obtained for SNLI. Given the difference between SNLI and Yelp (displayed in
Appendix \ref{EXAMPLESEXTRACTIONS}) there are two clear reasons for this decrease. The first is that Yelp is a dataset where it is
 harder to locate the 
syntactic roles. This is illustrated by the fact that the PB baseline obtains a much lower score. The second is that 
our syntactic role extraction heuristics are tailored for regular sentences with verbal roots, which subjects the evaluation metrics on Yelp 
to a considerable amount of noise. Nevertheless, the comparisons between a VAE, an ADVAE, and PB retain the same conclusions, but with lower margins and some 
overlapping standard deviations.

Through manual inspection of examples, we observed that the various structural characteristics (enumerations, sentences with nominal roots,
 presence of coordinating conjunctions, etc) were captured by different variables. This indicates that future iterations 
of our model need to provide ways to separate structural information from content-related information.

\section{Measuring the effect of latent variables on the structure of sentences}
\label{StructHeatMap}
\begin{figure*}[!h]
\centering
    \begin{minipage}[b]{0.48\textwidth}
            \centering
            \begin{minipage}[b]{\textwidth}
            \begin{adjustbox}{minipage=\textwidth,scale=0.5}
             \hspace{ 1cm} \includegraphics[trim={1.3cm 0.7cm 2.2cm 1.3cm},clip] {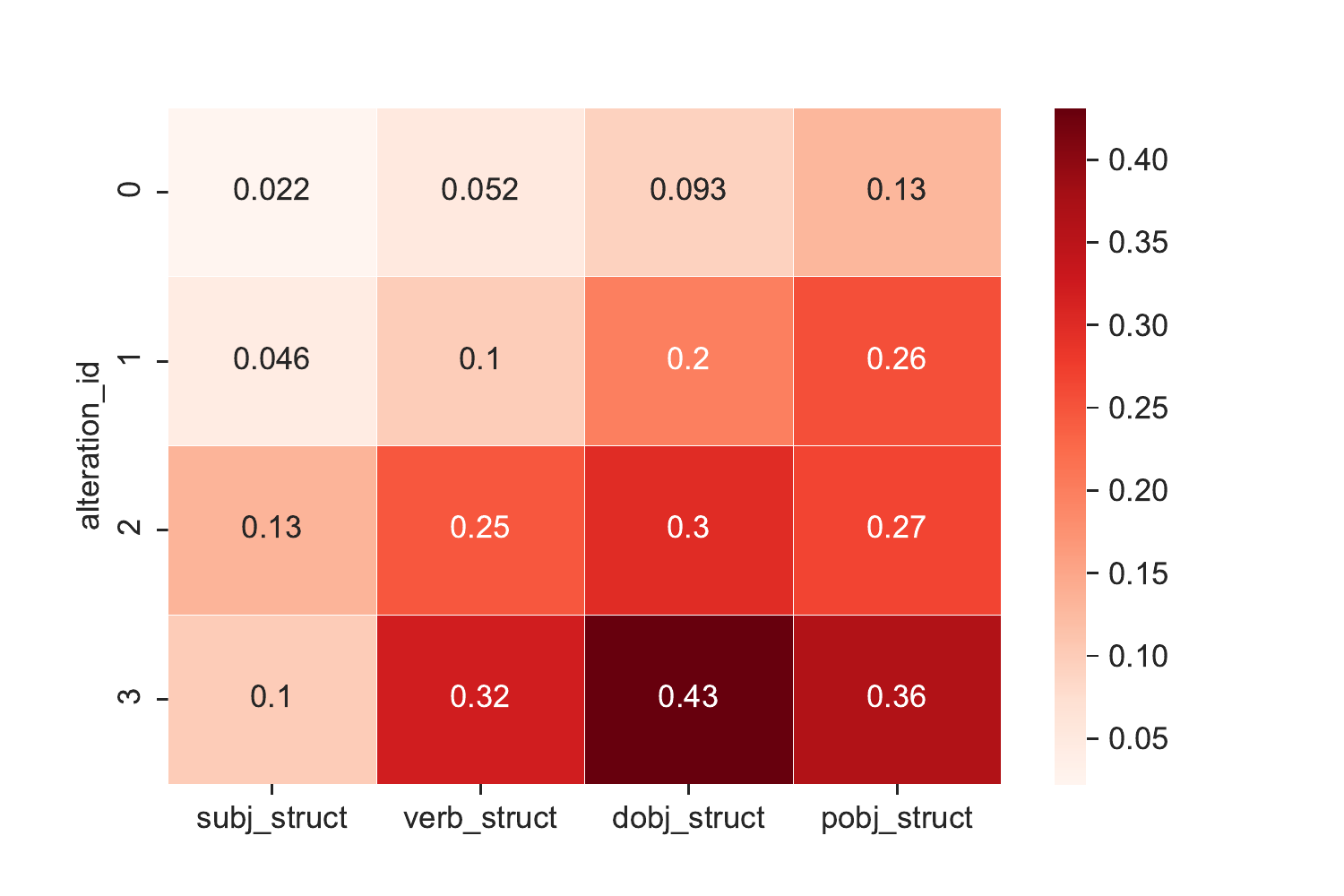}
            \end{adjustbox}
            \end{minipage}
            \caption{\centering The influence of latent variables on the appearance or disappearance of syntactic roles.}
            \label{fig:DECHEATStruct}
    \end{minipage}
\end{figure*}
In Figure \ref{fig:DECHEATStruct}, for each latent variable and each syntactic role, we report the probability that resampling the latent variable causes the appearance/disappearance of the syntactic role. The instance we use here is the same as the one we use for the heatmaps in the main body of the paper. According to the heatmaps in Figures \ref{fig:ENCHEAT} and \ref{fig:DECHEAT}, latent variable 3 is the one associated with the verb. As can be seen in the present heatmap in Figure \ref{fig:DECHEATStruct}, this same variable is the one that has the most influence on the appearance/disappearance of direct and prepositional objects, and this is a pattern that proved to be consistent across our different runs. This constitutes empirical justification  for our choice of discarding these cases from our decoder influence metrics. 

\section{Example Sentences from Yelp and SNLI and their Corresponding Syntactic Extractions}
\label{EXAMPLESEXTRACTIONS}
Table \ref{tab:EXAMPLESEXTRACTIONS} shows some samples from SNLI and Yelp reviews. Samples from 
Yelp Reviews exhibit a clearly higher structural diversity. On the other hand, most SNLI samples are highly similar in structure.\\ 
Our syntactic role extraction heuristics were tailored for sentences with verbal roots. As a result,
it can be seen that they struggle with sentences with nominal roots as well as other forms of irregular utterances present in Yelp. For SNLI, 
our extractions mostly yield the expected results, allowing for a reliable global assessment of our models.\\

\begin{table*}[!h]
    \small
    \centering
    \caption{Example syntactic role extractions from both SNLI and Yelp}
    \begin{tabularx}{14cm}{|c|p{4cm}|X|X|X|X|}
    \hline
     Source & Sentence & subj & verb & dobj & pobj \\
    \hline \hline
    Yelp &  i was originally told it would take \_num\_ mins . &  it& told& \_ num \_ mins&  \\\hline
    Yelp &  slow , over priced , i 'll go elsewhere next time . &  i& go& &  \\\hline
    Yelp &  we will not be back &  we& & &  \\\hline
    Yelp &  terrible . &  & & &  \\\hline
    Yelp &  at this point they were open and would be for another hour . &  they& & & this point \\\hline
    SNLI &  people are outside playing baseball . &  people& & baseball& \\\hline
    SNLI &  two dogs pull on opposite ends of a rope . &  two dogs& pull& opposite ends of a rope& a rope \\\hline
    SNLI &  a lady lays at a beach . &  a lady& lays& & a beach\\\hline
    SNLI &  people are running through the streets while people watch . &  people& running& & the streets   \\\hline
    SNLI &  someone prepares food into bowls  &  someone& prepares& food& bowls \\\hline
    \end{tabularx}
    \label{tab:EXAMPLESEXTRACTIONS}
\end{table*} 

\section{Training Details and Hyper-Parameter Settings}
\label{TRAINING&HP}
\paragraph{Our ADVAE's hyper-parameters}
Our model has been set to be large enough to reach a low reconstruction error during the initial reconstruction 
phase of the training. We use 2-layer Transformers with 4 attention heads and a hidden size of 192. Contrary to Vanilla VAEs, our model
seems to perform better with high values of $N_Z$. Therefore, we set our latent vector to a size of 768, and divide it into 96-dimensional
 variables for our $N_Z=8$ model and to 192-dimensional latent variables for our $N_Z=4$ model.
 No automated hyper-parameter selection has been done afterward. 

\paragraph{Sequence VAE hyper-parameters}
As is usually done for this baseline \citep{Xu2020OnSupervision}, we set both the encoder and the decoder to be 2-layer LSTMs.\\
We run this model for hidden LSTM sizes in [256, 512], and latent vector sizes in [16, 32]. The results for the model scoring the
 highest $\mathbb{D}_{dec}$ are then reported. Even though selection has been done according to $\mathbb{D}_{dec}$,
 we checked the remaining instances of our baselines and they also yielded low $N_{\Gammaopdec}$ values.\\
  
\paragraph{Transformer VAE hyper-parameters}
We set the hidden sizes and number of layers for this baseline similarly to ADVAE, since it is also a Transformer. We run this model for latent vector sizes in [16, 32] and display the highest scoring model, as is done for the Sequence VAE. 
\paragraph{Training phases}
All our models are trained using ADAM\citep{Kingma2015} with a batch size of 128 and a learning rate of 2e-4 for 20 epochs. The
dropout is set to 0.3.
To avoid posterior collapse, we train all our models for 3000 steps with $\beta=0$ (reconstruction phase), then we linearly increase 
$\beta$ to its final value for the subsequent 3000 steps. Following \citet{Bowman2016GeneratingSpace}, we also use
 word-dropout. We set its probability to 0.1.
\paragraph{Evaluation}
For the evaluation, $T^{dec}$ is set to 2000, and $T^{enc}$ is equal to the size of the test set.

\section{Disentanglement Scores for each Syntactic Role}
\label{FULLSYNRESULTS}
The full disentanglement scores are reported in Table \ref{tab:results1} for the decoder, and in Table \ref{tab:results2} for the encoder.
\begin{table*}[h]
    \centering
    \caption{Complete decoder disentanglement scores for SNLI}
    \resizebox{\textwidth}{!}{%
    \begin{tabular}{|c|c||c|c||c|c|c|c|}
    \hline
    Model& $\beta$ & $\mathbb{D}_{dec}$ &  $N_{\Gammaopdec}$  &  $\Delta\Gamma_{dec,verb}$ &  $\Delta\Gamma_{dec,subj}$ & $\Delta\Gamma_{dec,dobj}$ & $\Delta\Gamma_{dec,pobj}$ \\
    \hline \hline
    \multirow{3}{*}{ours-4} &0.3& 0.68\textcolor{gray}{(0.22)}& 2.80\textcolor{gray}{(0.45)}& 0.19\textcolor{gray}{(0.04)}& 0.35\textcolor{gray}{(0.18)}& 0.06\textcolor{gray}{(0.03)}& 0.07\textcolor{gray}{(0.03)}\\ 
     &0.4& 0.81\textcolor{gray}{(0.05)}& 3.00\textcolor{gray}{(0.00)}& 0.21\textcolor{gray}{(0.04)}& 0.47\textcolor{gray}{(0.03)}& 0.06\textcolor{gray}{(0.02)}& 0.07\textcolor{gray}{(0.02)}\\
    \hline
    \multirow{3}{*}{ours-8} &0.3& 0.60\textcolor{gray}{(0.10)}& 3.00\textcolor{gray}{(0.00)}& 0.17\textcolor{gray}{(0.04)}& 0.31\textcolor{gray}{(0.08)}& 0.05\textcolor{gray}{(0.04)}& 0.07\textcolor{gray}{(0.04)}\\ 
     &0.4& 0.63\textcolor{gray}{(0.35)}& 2.80\textcolor{gray}{(0.45)}& 0.17\textcolor{gray}{(0.10)}& 0.32\textcolor{gray}{(0.18)}& 0.05\textcolor{gray}{(0.04)}& 0.08\textcolor{gray}{(0.05)}\\

\hline
    \hline
    \multirow{3}{*}{Sequence VAE}&0.3& 0.60\textcolor{gray}{(0.09)}& 2.40\textcolor{gray}{(0.55)}& 0.24\textcolor{gray}{(0.06)}& 0.03\textcolor{gray}{(0.04)}& 0.03\textcolor{gray}{(0.02)}& 0.31\textcolor{gray}{(0.03)}\\ 
    &0.4& 1.28\textcolor{gray}{(0.24)}& 1.40\textcolor{gray}{(0.55)}& 0.45\textcolor{gray}{(0.12)}& 0.23\textcolor{gray}{(0.02)}& 0.02\textcolor{gray}{(0.02)}& 0.57\textcolor{gray}{(0.11)}\\
    \hline
    \multirow{3}{*}{Transformer VAE}&0.3& 0.12\textcolor{gray}{(0.10)}& 3.00\textcolor{gray}{(0.70)}& 0.01\textcolor{gray}{(0.01)}& 0.07\textcolor{gray}{(0.06)}& 0.01\textcolor{gray}{(0.01)}& 0.03\textcolor{gray}{(0.03)}\\ 
    &0.4& 0.11\textcolor{gray}{(0.04)}& 3.20\textcolor{gray}{(0.44)}& 0.03\textcolor{gray}{(0.02)}& 0.04\textcolor{gray}{(0.04)}& 0.01\textcolor{gray}{(0.01)}& 0.02\textcolor{gray}{(0.01)}\\
   
 \hline
     \end{tabular}}
    \label{tab:results1}
  \end{table*}
  \begin{table*}[h]
    \centering
    \caption{Complete encoder disentanglement scores for SNLI}
    \resizebox{\textwidth}{!}{%
    \begin{tabular}{|c|c||c|c||c|c|c|c|}
    \hline
    Model& $\beta$ & $\mathbb{D}_{enc}$ &  $N_{\Gammaopenc}$  &  $\Delta\Gamma_{enc,verb}$ &  $\Delta\Gamma_{enc,subj}$ & $\Delta\Gamma_{enc,dobj}$ & $\Delta\Gamma_{enc,pobj}$ \\
    \hline \hline
    \multirow{3}{*}{ours-4} &0.3& 1.30\textcolor{gray}{(0.09)}& 3.00\textcolor{gray}{(0.00)}& 0.28\textcolor{gray}{(0.05)}& 0.65\textcolor{gray}{(0.02)}& 0.08\textcolor{gray}{(0.03)}& 0.29\textcolor{gray}{(0.03)}\\ 
     &0.4& 1.46\textcolor{gray}{(0.33)}& 3.00\textcolor{gray}{(0.00)}& 0.38\textcolor{gray}{(0.12)}& 0.64\textcolor{gray}{(0.10)}& 0.14\textcolor{gray}{(0.04)}& 0.30\textcolor{gray}{(0.10)}\\
     \hline
    \multirow{3}{*}{ours-8} &0.3& 1.36\textcolor{gray}{(0.13)}& 3.40\textcolor{gray}{(0.89)}& 0.44\textcolor{gray}{(0.12)}& 0.60\textcolor{gray}{(0.18)}& 0.21\textcolor{gray}{(0.08)}& 0.11\textcolor{gray}{(0.06)}\\ 
     &0.4& 1.44\textcolor{gray}{(0.79)}& 3.40\textcolor{gray}{(0.55)}& 0.42\textcolor{gray}{(0.23)}& 0.61\textcolor{gray}{(0.34)}& 0.17\textcolor{gray}{(0.10)}& 0.23\textcolor{gray}{(0.16)}\\
     
\hline
    \hline
    \multirow{1}{*}{Average Position}& - & 0.98 \textcolor{gray}{(-)}& 3.00\textcolor{gray}{(-)}& 0.12\textcolor{gray}{(-)} & 0.70\textcolor{gray}{(-)} & 0.12\textcolor{gray}{(-)} & 0.04\textcolor{gray}{(-)}

\\ 
   
 \hline
     \end{tabular}}
    \label{tab:results2}
  \end{table*}

\section{Disentanglement Heatmaps Over the Entire Range of Syntactic Roles and PoS Tags}
\label{ENTIREROLERESULTS}

\begin{figure*}[!h]
    \centering
    \begin{minipage}[b]{\textwidth}
    \begin{adjustbox}{minipage=\textwidth,scale=0.3}
    \includegraphics[trim={8.9cm 0.7cm 19cm 1.3cm},clip] {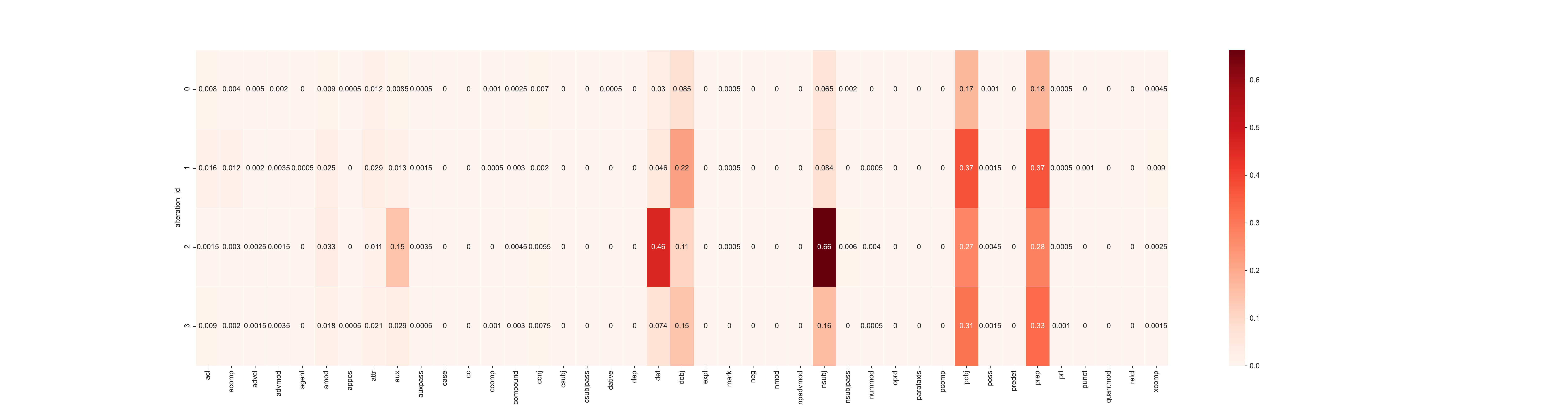}
    \end{adjustbox}
    \end{minipage}
    \caption{Decoder influence heatmap for all SD syntactic roles.}
    \label{fig:ENCHEATLDC}
\end{figure*}

\begin{figure*}[!h]
    \centering
    \begin{minipage}[b]{\textwidth}
    \begin{adjustbox}{minipage=\textwidth,scale=0.3}
    \includegraphics[trim={8.9cm 0.7cm 19cm 1.3cm},clip] {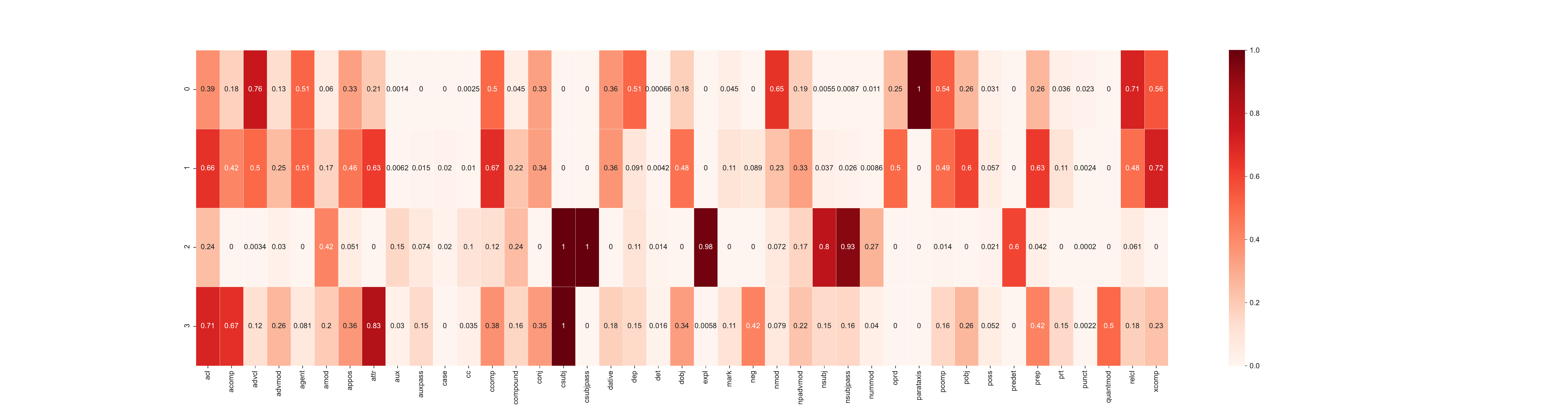}
    \end{adjustbox}
    \end{minipage}
    \caption{Encoder influence heatmap for all SD syntactic roles.}
    \label{fig:DECHEATLDC}
\end{figure*}

\begin{figure*}[!h]
    \centering
    \begin{minipage}[b]{\textwidth}
    \begin{adjustbox}{minipage=\textwidth,scale=0.33}
    \includegraphics[trim={8.9cm 0.7cm 25cm 1.3cm},clip] {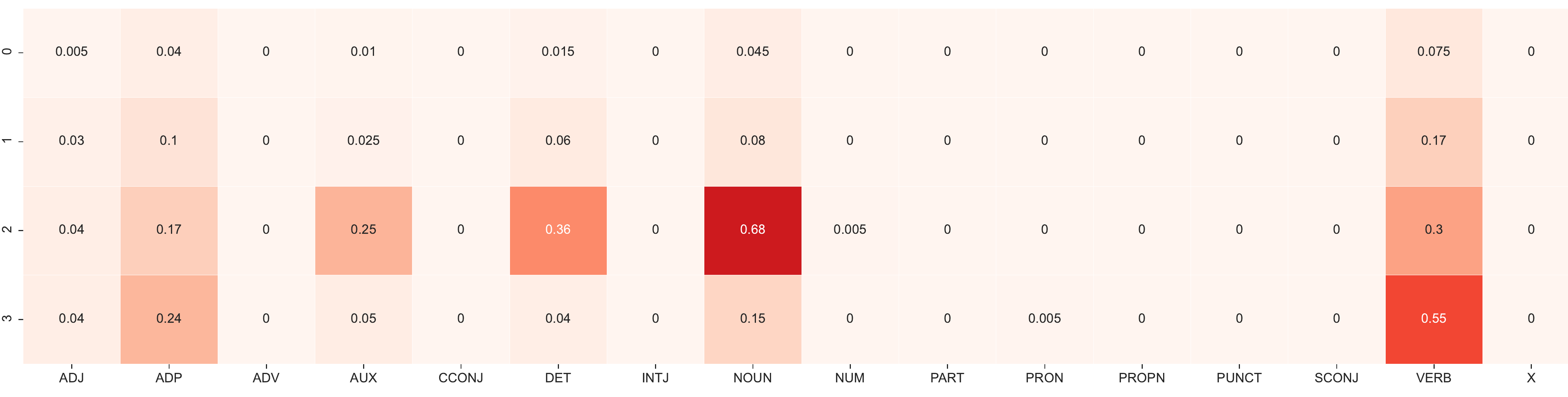}
    \end{adjustbox}
    \end{minipage}
    \caption{Decoder influence heatmap for all PoS Tags.}
    \label{fig:ENCHEATPOS}
\end{figure*}

\begin{figure*}[!h]
    \centering
    \begin{minipage}[b]{\textwidth}
    \begin{adjustbox}{minipage=\textwidth,scale=0.3}
    \includegraphics[trim={8.9cm 0.7cm 19cm 1.3cm},clip] {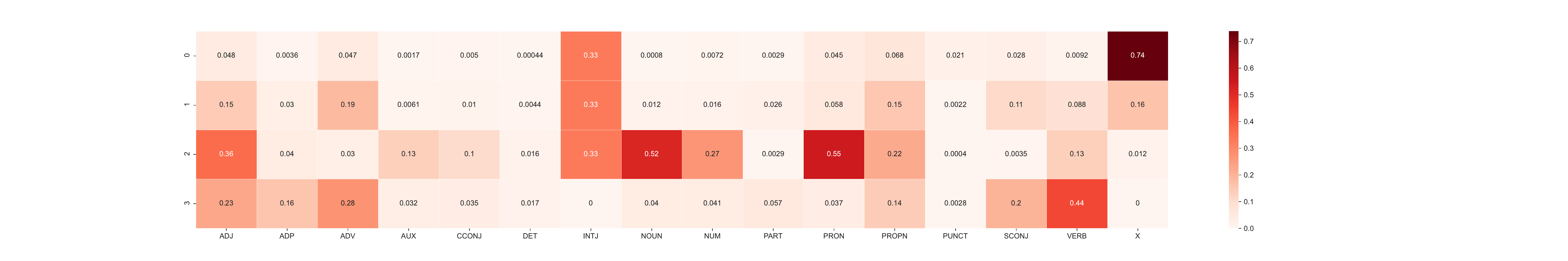}
    \end{adjustbox}
    \end{minipage}
    \caption{Encoder influence heatmap for all PoS Tags.}
    \label{fig:DECHEATPOS}
\end{figure*}

\begin{figure*}[!h]
    \centering
    \begin{minipage}[b]{\textwidth}
    \begin{adjustbox}{minipage=\textwidth,scale=0.3}
    \includegraphics[trim={8.9cm 0.7cm 19cm 1.3cm},clip] {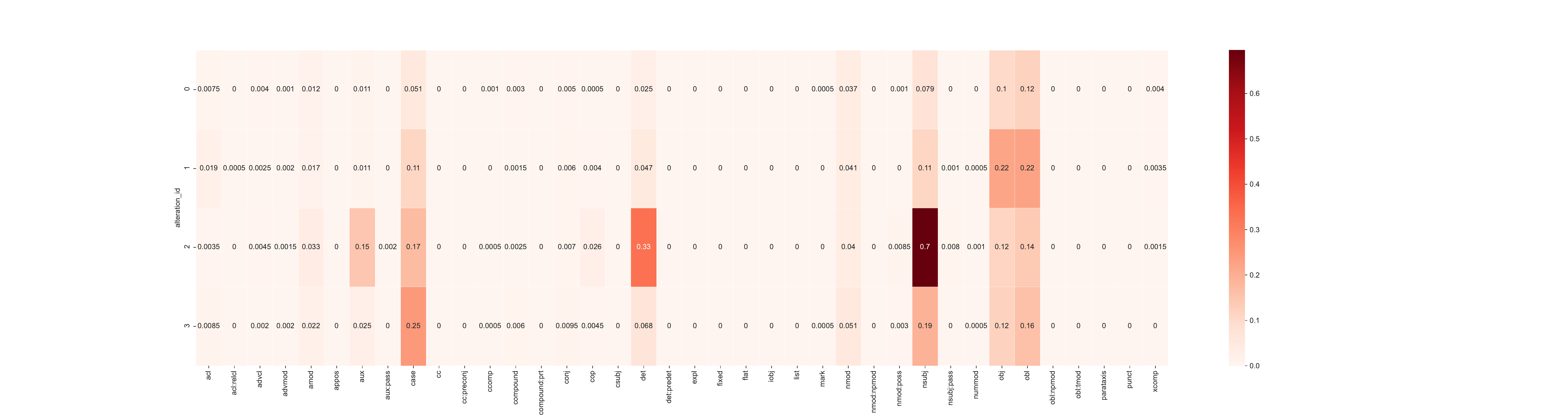}
    \end{adjustbox}
    \end{minipage}
    \caption{Decoder influence heatmap for all UD syntactic Roles.}
    \label{fig:ENCHEATUD}
\end{figure*}

\begin{figure*}[!h]
    \centering
    \begin{minipage}[b]{\textwidth}
    \begin{adjustbox}{minipage=\textwidth,scale=0.3}
    \includegraphics[trim={8.9cm 0.7cm 19cm 1.3cm},clip] {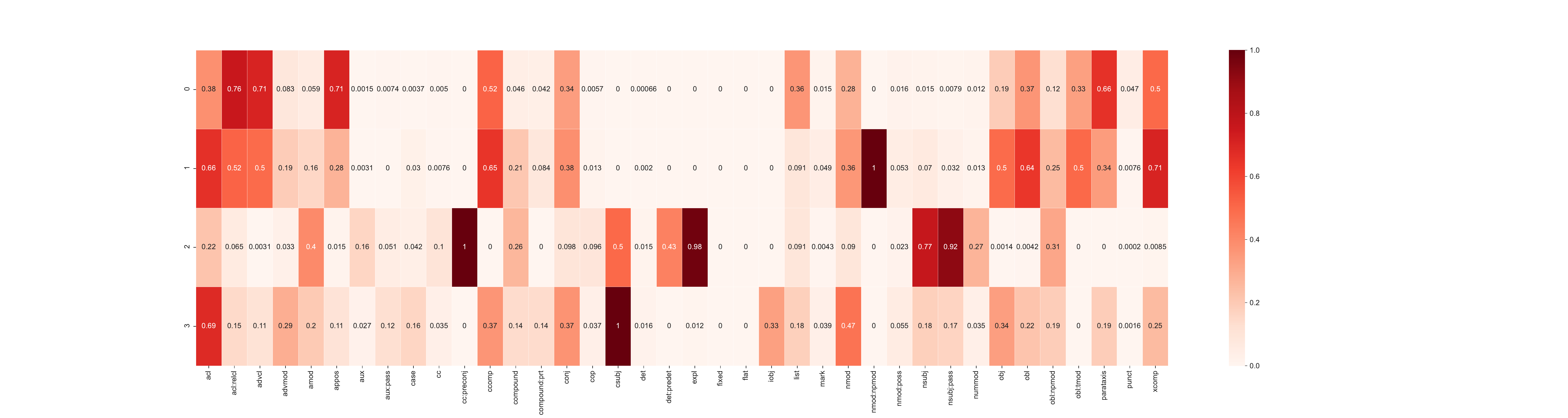}
    \end{adjustbox}
    \end{minipage}
    \caption{Encoder influence heatmap for all UD syntactic Roles.}
    \label{fig:DECHEATUD}
\end{figure*}

We report decoder and encoder heatmaps for all the syntactic roles following the Stanford Dependencies (SD; \citealp{de2008stanford}) annotation scheme of Ontonotes, which was used to train our Spacy2 parser, in Figures \ref{fig:ENCHEATLDC} and \ref{fig:DECHEATLDC}. For the sake of extensiveness and to make sure we did not draw results from some parser biases, we also report the same heatmaps but using UDPipe 2.0 \citep{straka-2018-udpipe}, which uses UD type annotations\footnote{A widely adopted annotation scheme derived from Stanford Dependencies.}, in Figures  \ref{fig:ENCHEATUD} and \ref{fig:DECHEATUD}. Finally, we also report heatmaps for interaction with PoS Tags extracted with Spacy2 in Figures  \ref{fig:ENCHEATPOS} and \ref{fig:DECHEATPOS}. As was done in the main body of the paper, the span corresponding to each syntactic role (in both annotation schemes) was taken to be the series of words included in its corresponding subtree. In contrast, the span corresponding to each PoS tag was just taken to be the tagged word. Results from UD parsing extraction lead to the same conclusions as from our initial SD results.

The instance of our ADVAE for which we display the above heatmaps is the same one for which we display the heatmaps in Figures \ref{fig:ENCHEAT} and \ref{fig:DECHEAT} in the main body of the paper. As shown in those Figures, it mostly uses variable 3 for verbs, variables 2 for subjects, and variable 1 for objects. The remaining variable (0) also seems to capture some interaction with objects.  The heatmaps show that our ADVAE tends to group syntactic roles into latent variables in a way that aligns with the predicative structure of sentences. In fact, variable 2 displays the highest influence on the PoS tag VERB as well as its surroundings as a predicate argument such as adverbs and adverbial phrases. Similarly, latent variable 2 displays a high influence on subjects (nominal or clausal), numeral modifiers, adjectival modifiers, and auxiliaries (for conjugation). Moreover, Variable 1 highly influences the direct and prepositional objects, which we study in the main body of the paper, but also diverse clausal modifiers and obliques which often play similar roles to direct and prepositional objects in a predicate structure.

\section{Additional Examples of Resampled realizations for each syntactic role}
\label{QUALIAPPEN}
Table \ref{tab:resultsresamplebig} contains a wide array of examples where the latent variable corresponding to each syntactic role is resampled.
\begin{table}[h!]
    \small
    \centering
    \caption{More examples where we resample a specific latent variable for a sentence.}
    \begin{tabularx}{14cm}{|X|X|X|X|}
    \hline
     Original sentence& Resampled subject& Resampled verb & Resampled dobj/pobj \\
    \hline \hline
     the woman is riding a large brown dog  & two men are riding in a large city  & the woman is wet  & the woman is riding on the bus \\ \hline
 the police are running in a strategy  & a man is looking at a date  & the police are at an arid  & the police are running in a wooded area \\ \hline
 a man is holding a ball  & a man is holding a ball  & a man is , and a woman are talking on a road  & a man is sitting on a cellphone outside \\ \hline
 everyone is watching the game  & some individuals are watching tv  & everyone is a man  & everyone is watching the game in the air \\ \hline
 there is a man in the air  & a man is sitting in the air  & there is no women wearing swim trunks  & there is a man in a red shirt \\ \hline
 a group of friends are standing on a beach  & an elderly father and child are standing on the beach  & a group of people are standing on a beach  & a group of friends are looking at the beach \\ \hline
 the women are in a store  & a man is playing a game  & two women are on a break  & two women are sitting on a bench \\ \hline
 a man is playing a game  & a little girl is playing with a ball  & a man is clean  & a man is sitting on a lake to an old country \\ \hline
 a man is playing a game  & some dogs are playing in the pool  & a man is preparing to chase himself  & a man is playing a game \\ \hline
 the memorial woman is happy  & a dog is happy  & the memorial workers are in a room  & the memorial is happy \\ \hline
 a man is wearing a green jacket and a ship  & a boy sitting in a green device  & a man is dancing for the camera  & the man is wearing a hat \\ \hline
 a man is playing a game  & a man is playing a game  & two men are tripod  & a man is playing with a guitar \\ \hline
 a man is wearing a brown sweater and green shirt  & a karate dog is swimming in a chair  & a man is bought a brown cat in an airplane  & a man is wearing a dress and talks to the woman \\ \hline
 the woman is about to visitors  & three people are working at a babies  & the woman is wearing a sewer  & the woman is about to sell a tree \\ \hline
 a man is sitting in the snowy field  & a man is sitting in the snowy field  & a man is wearing electronics  & a man is sitting on a park bench \\ \hline
 two people are playing in the snow  & the motorcycle is a woman on the floor  & two people play soccer in the snow  & two people are playing in a concert \\ \hline
 a man is standing next to another man  & a boy is standing next to another man  & a man is standing  & a man is standing next to a man \\ \hline
 a man is on his bike  & a man is on his bike  & a dog is showing water  & a man is on his bike \\ \hline
 a man is sitting in front of a tree , taking a picture  & a man is sitting in front of a tree  & a man is holding a red shirt and climbing a tree  & a man is sitting on a suburban own \\ \hline
 a man is sitting with a dog  & the children are sitting with the dog  & a man is playing with a dog  & a man is sitting with an umbrella \\ \hline
 the man is holding a ball  & a boy is playing with a ball  & the man is on a bike  & the man is waiting for a counts to jump for the first base \\ \hline
 a man is holding a game  & five people buying a skateboard from easter  & a man is on a bicycle  & a man is very large \\ \hline
 two men are playing in a field  & the kids play in the snow  & two men are playing a game  & two men are playing in a field \\ \hline
 a man is wearing a hat  & a woman is wearing a hat  & the man is they oil  & a man is wearing a black bathing suit near buildings \\ \hline
 a man is playing a guitar  & the man is wearing a blue shirt  & a man is sitting on a bench  & a man wearing a hat is playing a guitar \\ \hline
 a woman is playing a game  & a man is playing a game  & a woman is playing a game  & a woman is playing a game \\ \hline
 a man is on the truck  & the people are on the truck  & a man is holding a truck  & a man is on the grass \\ \hline
 a man is playing with the cut  & a small boy is playing on the cut  & a man is warming up the cut  & a man is playing a game \\ \hline
 a group of people are at a park  & the man is wearing a blue shirt  & a group of people are at a park  & a group of people are at a park \\ \hline
     \end{tabularx}
    \label{tab:resultsresamplebig}
\end{table} 

\clearpage
\section{Reconstruction and Kullback-Leibler Values Across Experiments}

\begin{table}[!h]
    \caption{Reconstruction loss and Kullback-Leibler values on SNLI.}
    \centering
    \resizebox{0.8\textwidth}{!}{%
    \begin{tabular}{|c|c||c|c|c|}
    \hline
    Model& $\beta$ & $-\mathbb{E}_{(z) \sim q_\phi(z|x)}\left[ \log p_\theta(x|z) \right] $& $\KL[q_\phi(z|x)||p(z)]$& Perplexity Upper Bound\\
    \hline \hline
    
    \multirow{2}{*}{Sequence VAE}&0.3&  31.38\textcolor{gray}{(0.12)}& 2.80\textcolor{gray}{(0.25)}&22.02\textcolor{gray}{(0.30)}\\ 
    &0.4& 32.19\textcolor{gray}{(0.13)}& 1.22\textcolor{gray}{(0.04)}&21.08\textcolor{gray}{(0.22)}\\
    \hline
    
    \multirow{2}{*}{Transformer VAE}&0.3&  24.35\textcolor{gray}{(0.14)}& 13.38\textcolor{gray}{(0.19)}&25.07\textcolor{gray}{(0.27)}\\ 
    &0.4& 26.57\textcolor{gray}{(0.27)}& 8.36\textcolor{gray}{(0.32)}&20.68\textcolor{gray}{(0.16)}\\
    \hline
    \hline
    \multirow{2}{*}{ours-4} &0.3&  10.75\textcolor{gray}{(0.94)}& 42.63\textcolor{gray}{(1.16)}&68.49\textcolor{gray}{(5.96)}\\ 
    &0.4& 16.01\textcolor{gray}{(0.64)}& 27.93\textcolor{gray}{(1.52)}&36.16\textcolor{gray}{(2.20)}\\
    \hline
    \multirow{2}{*}{ours-8} &0.3& 8.83\textcolor{gray}{(1.66)}& 46.99\textcolor{gray}{(2.99)}&77.26\textcolor{gray}{(9.02)}\\ 
     &0.4& 16.84\textcolor{gray}{(8.50)}& 27.34\textcolor{gray}{(14.99)}&39.23\textcolor{gray}{(11.27)}\\

 \hline
     \end{tabular}}
    \label{tab:resultsRecKL}
  \end{table}
\label{RECKL}

\begin{table}[!h]
    \caption{Reconstruction loss and Kullback-Leibler values on Yelp.}
    \centering
    \resizebox{0.8\textwidth}{!}{%
    \begin{tabular}{|c|c||c|c|c|}
    \hline
    Model& $\beta$ & $-\mathbb{E}_{(z) \sim q_\phi(z|x)}\left[ \log p_\theta(x|z) \right] $& $\KL[q_\phi(z|x)||p(z)]$& Perplexity Upper Bound\\
    \hline \hline
    \multirow{2}{*}{Sequence VAE}&0.3&  32.55\textcolor{gray}{(0.27)}& 4.26\textcolor{gray}{(0.57)}&36.97\textcolor{gray}{(0.82)}\\ 
    
    &0.4& 33.35\textcolor{gray}{(0.11)}& 1.42\textcolor{gray}{(0.15)}&32.42\textcolor{gray}{(0.13)}\\
    \hline
    \multirow{2}{*}{Transformer VAE}&0.3&  23.64\textcolor{gray}{(0.11)}&
    19.24\textcolor{gray}{(0.32)}&53.94\textcolor{gray}{(1.14)}\\ 
    &0.4& 26.41\textcolor{gray}{(0.11)}& 12.79\textcolor{gray}{(0.20)}&41.25\textcolor{gray}{(0.55)}\\
    \hline
    \hline
    \multirow{2}{*}{ours-4} &0.3&  7.30\textcolor{gray}{(0.27)}& 55.19\textcolor{gray}{(0.30)}&121.44\textcolor{gray}{(5.16)}\\ 
    &0.4& 18.19\textcolor{gray}{(4.38)}& 29.85\textcolor{gray}{(6.52)}&58.11\textcolor{gray}{(8.40)}\\
    \hline
    \multirow{2}{*}{ours-8} &0.3& 5.36\textcolor{gray}{(0.48)}& 59.24\textcolor{gray}{(0.61)}&129.54\textcolor{gray}{(7.63)}\\ 
     &0.4& 15.36\textcolor{gray}{(5.75)}& 34.16\textcolor{gray}{(12.40)}&63.62\textcolor{gray}{(16.56)}\\

 \hline
     \end{tabular}}
    \label{tab:resultsRecKLYelp}
  \end{table}
\label{RECKL}
The values for the reconstruction loss, the $\KL$ divergence, and the upper bound on perplexity concerning the experiments in the main body of the paper are reported in Table \ref{tab:resultsRecKL}. YThe same value for the Yelp experiments are in Table \ref{tab:resultsRecKLYelp}. Since our models are VAE-based, one can only obtain the upper bound on the perplexity and not its exact value. These upper bound values are obtained using an importance sampling-based estimate of the negative log-likelihood, as was done in \cite{Wu2020OnBeyond}. We set the number of importance samples to 10.

It can be seen that the behavior of ADVAEs is very different from classical Sequence VAEs and Transformer VAEs. On the plus side, they are capable of sustaining much more information in their latent variables as shown by their higher $\KL$, and they do better at reconstruction. The upper bound estimate of their perplexity is however higher. A high $\KL$ makes it more difficult for the importance sampling-based perplexity estimate to reach the true value of the model's perplexity. This may be the reason behind the higher values observed for ADVAEs. 

\section{Layer-wise Encoder Attention}
\label{PerLayerAtt}
\begin{figure*}[!h]
\centering
    \begin{minipage}[b]{0.30\textwidth}
            \centering
           \begin{minipage}[b]{\textwidth}
            \begin{adjustbox}{minipage=\textwidth,scale=0.35}
            \hspace{ 1cm} \includegraphics[trim={1.3cm 0.7cm 2.2cm 1.3cm},clip] {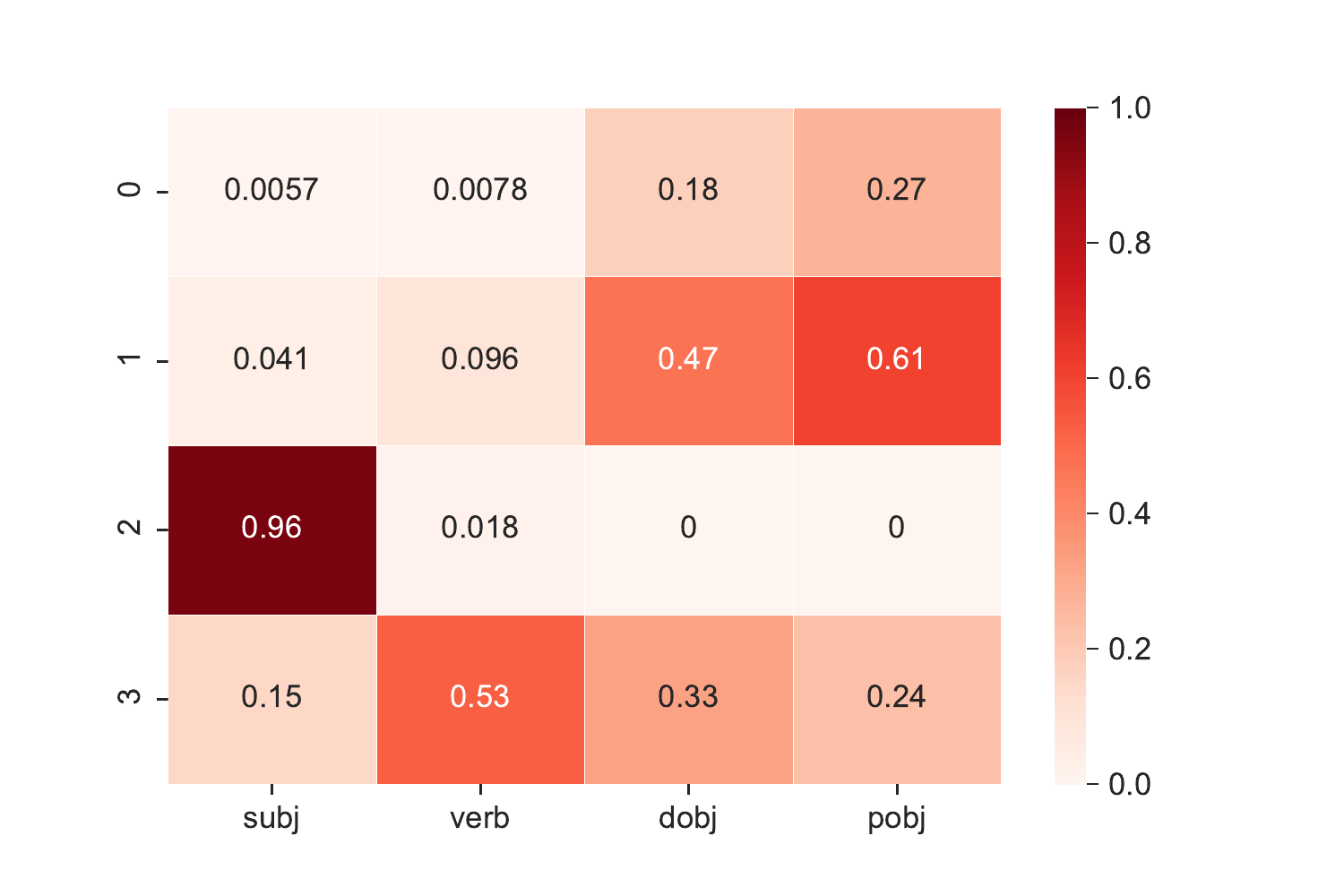}
            \end{adjustbox}
            \end{minipage}
            \caption{\centering Encoder influence heatmap ($\Gammaopenc$) when only using the \emph{first} layer.}
            \label{fig:ENCHEATLAYER0}
    \end{minipage}
    \begin{minipage}[b]{0.30\textwidth}
            \centering
            \begin{minipage}[b]{\textwidth}
            \begin{adjustbox}{minipage=\textwidth,scale=0.35}
             \hspace{ 1cm} \includegraphics[trim={1.3cm 0.7cm 2.2cm 1.3cm},clip] {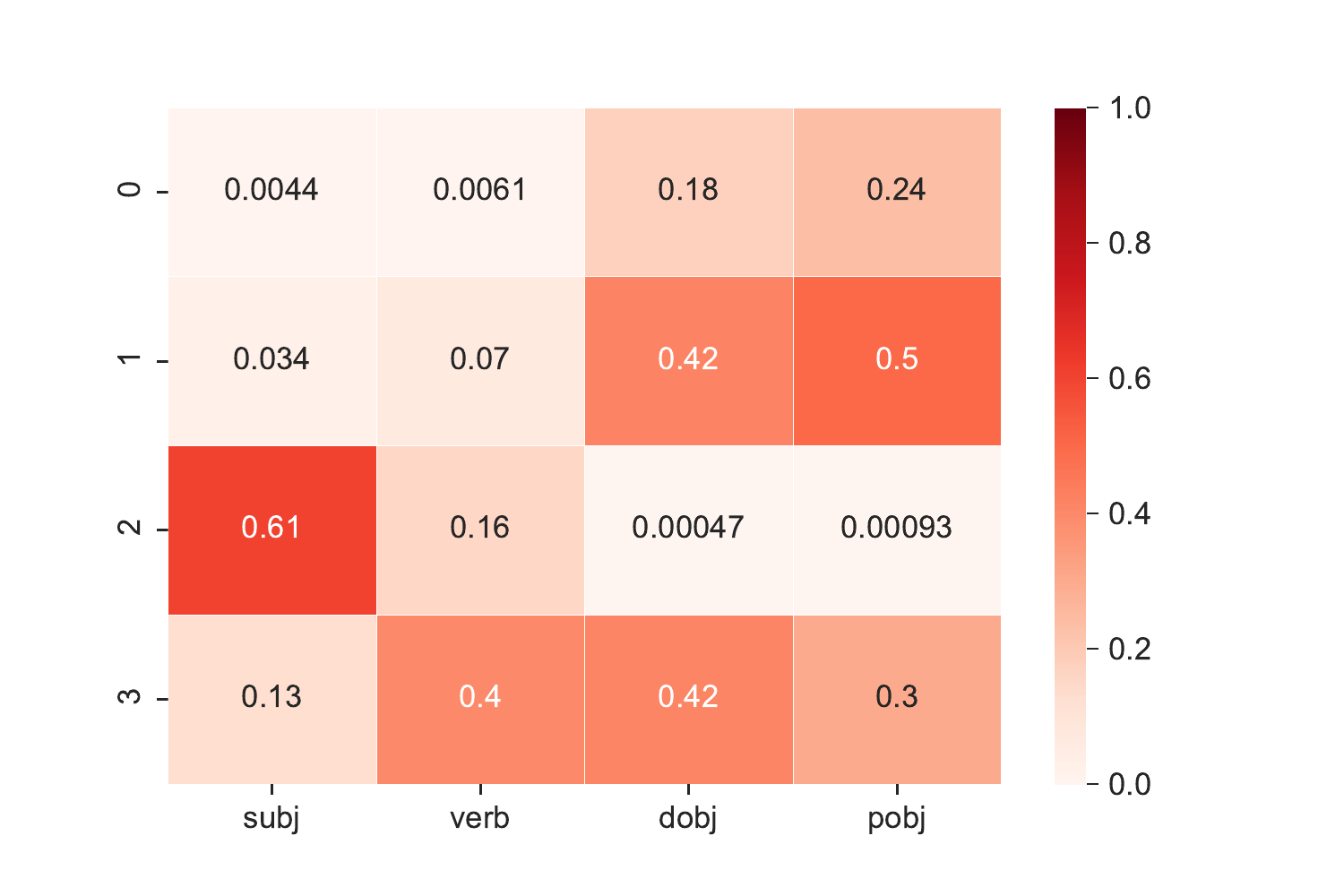}
            \end{adjustbox}
            \end{minipage}
            \caption{\centering Encoder influence heatmap ($\Gammaopenc$) when only using the \emph{second} layer.}
            \label{fig:ENCHEATLAYER1}
    \end{minipage}
    \begin{minipage}[b]{0.30\textwidth}
            \centering
           \begin{minipage}[b]{\textwidth}
            \begin{adjustbox}{minipage=\textwidth,scale=0.35}
            \hspace{ 1cm} \includegraphics[trim={1.3cm 0.7cm 2.2cm 1.3cm},clip] {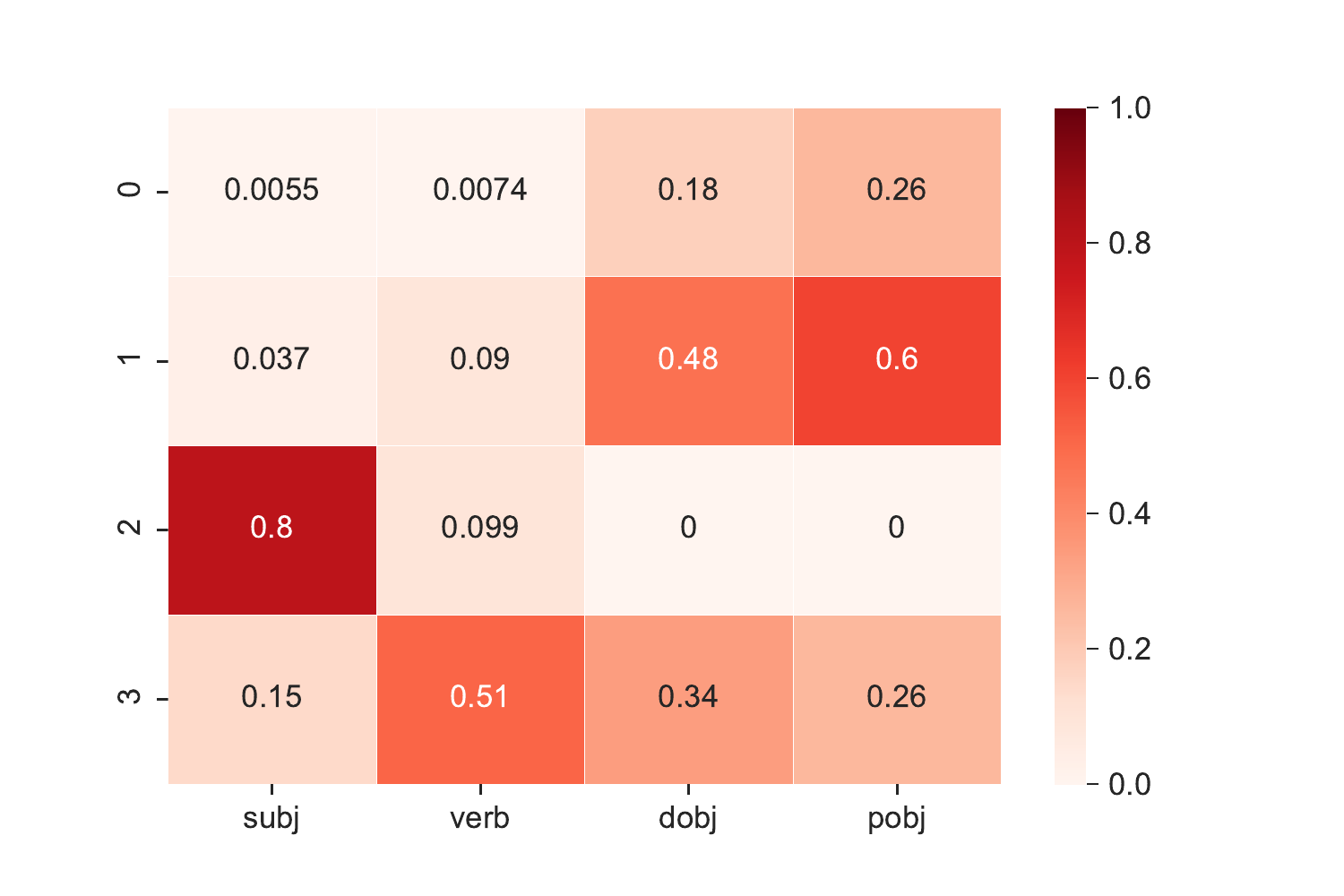}
            \end{adjustbox}
            \end{minipage}
            \caption{\centering Encoder influence heatmap ($\Gammaopenc$) when \emph{averaging} over both layers.}
            \label{fig:ENCHEATLAYERMEAN}
    \end{minipage}
\end{figure*}

In the main body of the paper, we use attention values that are averaged throughout the network. We hereby display the encoder heatmaps obtained by using attention values from the first layer (Fig. \ref{fig:ENCHEATLAYER0}), the second layer (Fig. \ref{fig:ENCHEATLAYER1}), or an average on both layers (Fig. \ref{fig:ENCHEATLAYERMEAN}) for comparison.

As can be seen, the first layer alone provides the most sparse heatmap, and thus, the clearest correspondence between syntactic roles and latent variables. This corroborates the claims of \cite{Tenney2020BERTPipeline} about syntax being most prominently processed in the early layers of Transformers.

\section{ADVAE Results for a larger grid of $N_z$ values}
\label{NZVARY}

We display in Table \ref{tab:NzVAry} the quantitative results of ADVAE on SNLI for $N_z$ in $\{2, 4, 6, 8\}$.  For ours-2, it is normal that it only separates syntactic role realizations into a maximum of 2 latent variables, as seen from the values of $N_{\Gammaopenc}$ and $N_{\Gammaopenc}$ , since 2 is its total number of latent variables.

\begin{figure*}[!h]
\hspace{-1cm}
    \begin{minipage}[b]{0.48\textwidth}
            \centering
           \begin{minipage}[b]{\textwidth}
            \begin{adjustbox}{minipage=\textwidth,scale=0.6}
            \hspace{ 1cm} \includegraphics[trim={1.3cm 0.7cm 2.2cm 1.3cm},clip] {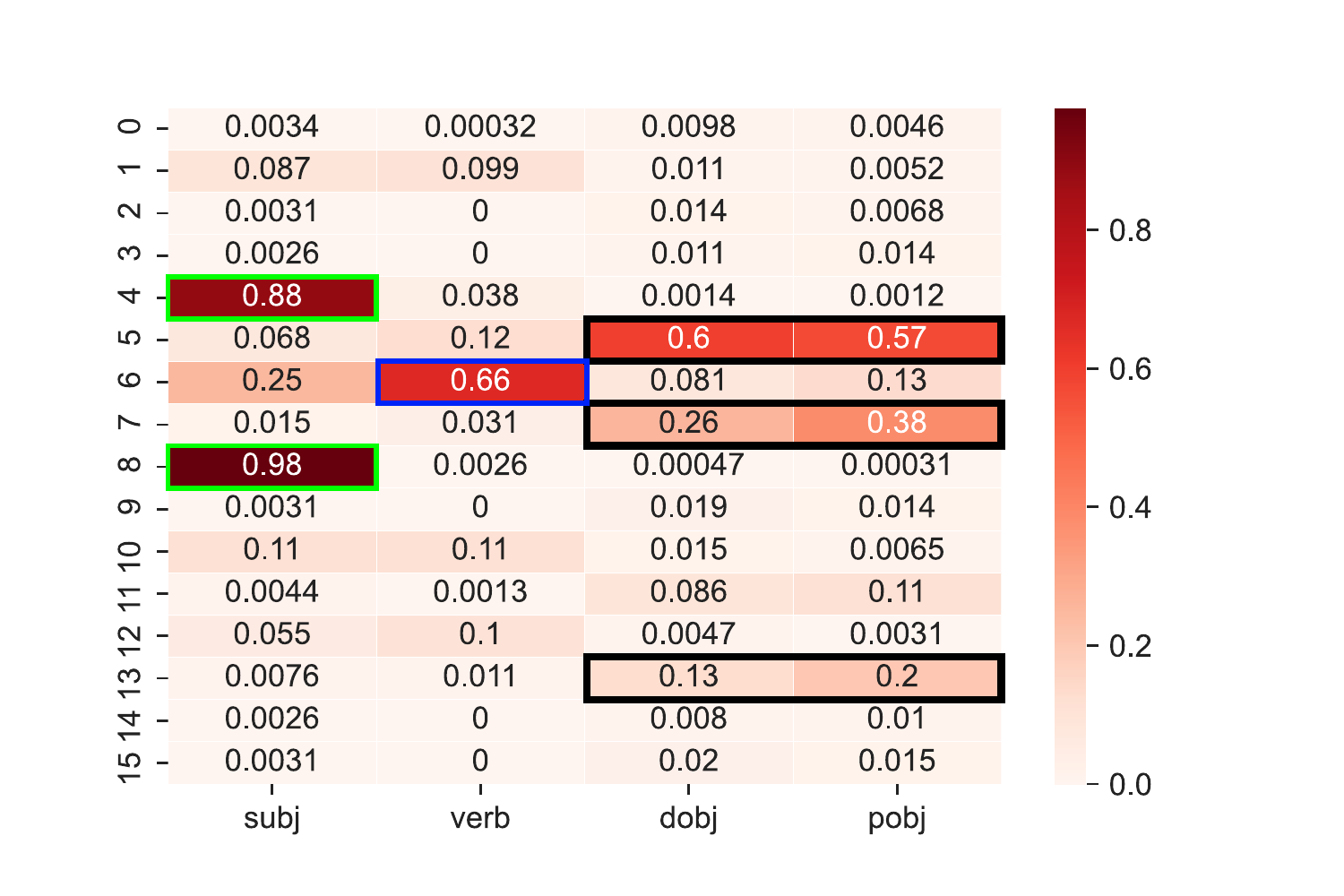}
            \end{adjustbox}
            \end{minipage}
            \caption{\centering Encoder influence heatmap for ADVAE with 16 latent variables on SNLI ($\Gammaopenc$). Squares with similar colors highlight groups of latent variables that relate to the same syntactic role.}
            \label{fig:ENCHEAT16}
    \end{minipage}
\hspace{0.7cm}
    \begin{minipage}[b]{0.48\textwidth}
            \centering
            \begin{minipage}[b]{\textwidth}
            \begin{adjustbox}{minipage=\textwidth,scale=0.6}
             \hspace{ 1cm} \includegraphics[trim={1.3cm 0.7cm 2.2cm 1.3cm},clip] {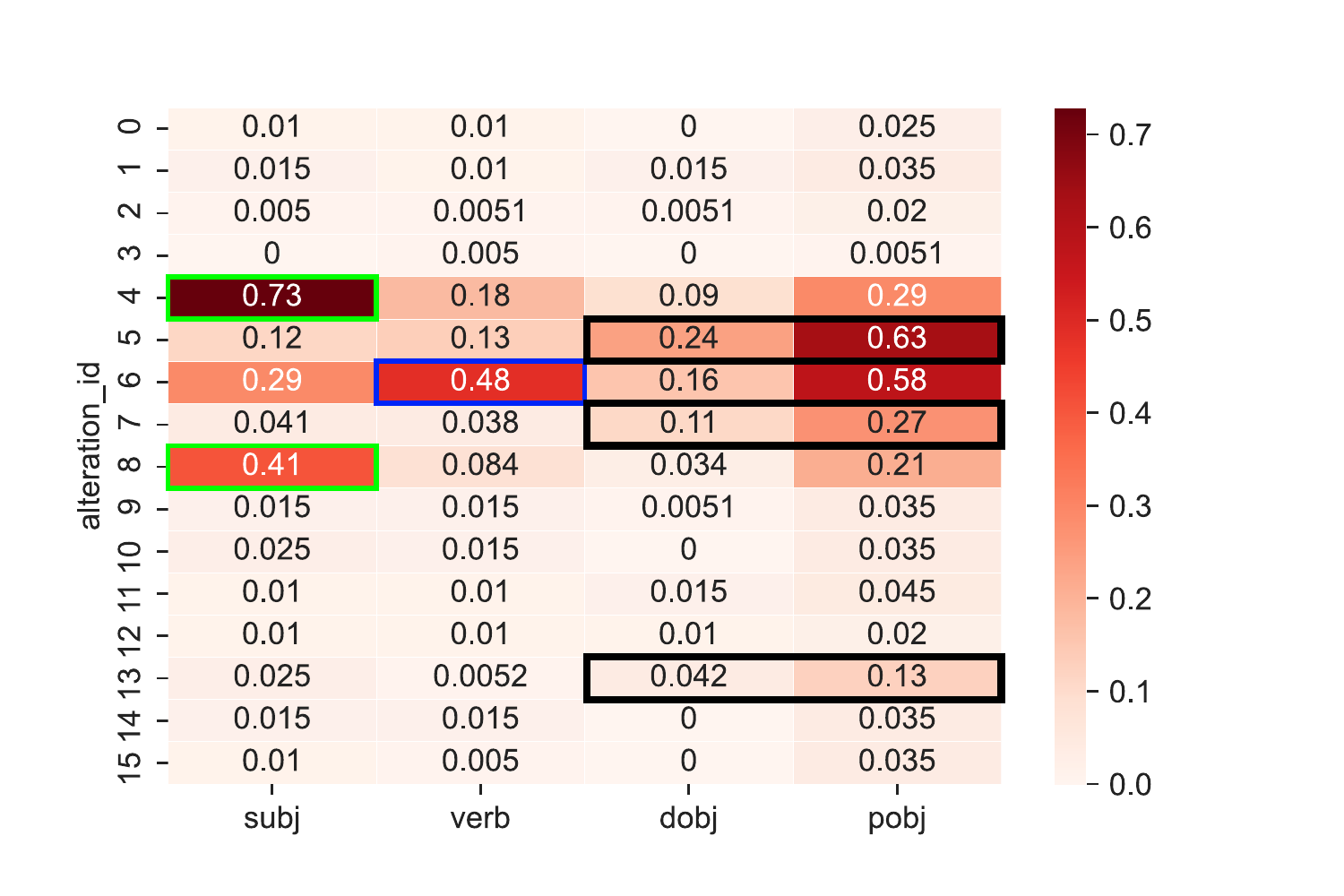}
            \end{adjustbox}
            \end{minipage}
            \caption{\centering Decoder influence heatmap for ADVAE with 16 latent variables on SNLI ($\Gammaopdec$). Squares with similar colors highlight groups of latent variables that relate to the same syntactic role. }
            \label{fig:DECHEAT16}
    \end{minipage}
\end{figure*}

As observed in the main body of the paper, the increase of the number of latent variables used in ADVAE leads to dispatching the influence on the realization of a single syntactic role to multiple latent variables. This in turn,leads to for $\mathbb{D}_{enc}$ and $\mathbb{D}_{dec}$ that we observe . In Figures \ref{fig:ENCHEAT16} and  \ref{fig:DECHEAT16}, we respectively display the encoder and decoder heatmaps of ADVAE with 16 latent variables. As can be seen in these figures, each latent variable still highly specializes in a specific syntactic role. This is seen more clearly on the encoder heatmap due to co-adaptation harming the clarity of the decoder heatmap. This specialization seems to be shared among groups (\textit{e.g.} variables 4 and 8 specialize in the subject, as indicated by the green squares on the figure). This causes the difference of influence between the most influential  variable and the second most influential one to be low, and thus decreases the values of $\mathbb{D}_{enc}$ and $\mathbb{D}_{dec}$.
\begin{table}[!h]
    \centering
    \caption{Disentanglement quantitative results on SNLI for a larger grid of $N_z$ values.}
    \resizebox{0.7\textwidth}{!}{%
    \begin{tabular}{|c|c||c|c||c|c|}
    \hline
    Model& $\beta$ & $\mathbb{D}_{enc}$ &  $N_{\Gammaopenc}$  &  $\mathbb{D}_{dec}$ &  $N_{\Gammaopdec}$\\
    \hline \hline
    
    \multirow{2}{*}{ours-2} &0.3& 2.01\textcolor{gray}{(0.07)}& 2.00\textcolor{gray}{(0.00)}& 0.92\textcolor{gray}{(0.06)}& 2.00\textcolor{gray}{(0.00)}\\ 
     &0.4& 0.33\textcolor{gray}{(0.15)}& 1.60\textcolor{gray}{(0.55)}& 0.13\textcolor{gray}{(0.09)}& 1.20\textcolor{gray}{(0.45)}\\
    \hline
    \multirow{2}{*}{ours-4} &0.3& 1.30\textcolor{gray}{(0.09)}& 3.00\textcolor{gray}{(0.00)}& 0.68\textcolor{gray}{(0.22)}& 2.80\textcolor{gray}{(0.45)}\\ 
     &0.4& 1.46\textcolor{gray}{(0.33)}& 3.00\textcolor{gray}{(0.00)}& 0.81\textcolor{gray}{(0.05)}& 3.00\textcolor{gray}{(0.00)}\\
    \hline
    \multirow{2}{*}{ours-8} &0.3& 1.36\textcolor{gray}{(0.13)}& 3.40\textcolor{gray}{(0.89)}& 0.60\textcolor{gray}{(0.10)}& 3.00\textcolor{gray}{(0.00)}\\ 
     &0.4& 1.44\textcolor{gray}{(0.79)}& 3.40\textcolor{gray}{(0.55)}& 0.63\textcolor{gray}{(0.35)}& 2.80\textcolor{gray}{(0.45)}\\
    \hline
    \multirow{2}{*}{ours-16} &0.3& 0.60\textcolor{gray}{(0.31)}& 3.60\textcolor{gray}{(0.55)}& 0.33\textcolor{gray}{(0.30)}& 2.60\textcolor{gray}{(0.55)}\\ 
     &0.4& 0.65\textcolor{gray}{(0.16)}& 3.40\textcolor{gray}{(0.55)}& 0.56\textcolor{gray}{(0.28)}& 2.60\textcolor{gray}{(0.55)}\\
    \hline

 \hline
     \end{tabular}}
    \label{tab:NzVAry}
  \end{table}

\end{document}

%% file: math_commands.tex
\usepackage{amsmath,amsfonts,bm}

\def\eqref#1{equation~\ref{#1}}

\def\1{\bm{1}}

\DeclareMathAlphabet{\mathsfit}{\encodingdefault}{\sfdefault}{m}{sl}
\SetMathAlphabet{\mathsfit}{bold}{\encodingdefault}{\sfdefault}{bx}{n}

\newcommand{\E}{\mathbb{E}}

\newcommand{\R}{\mathbb{R}}

\newcommand{\KL}{D_{\mathrm{KL}}}

\DeclareMathOperator*{\argmax}{arg\,max}

%% file: draft_final.tex
\expandafter\def\expandafter\UrlBreaks\expandafter{\UrlBreaks
  \do\a\do\b\do\c\do\d\do\e\do\f\do\g\do\h\do\i\do\j%
  \do\k\do\l\do\m\do\n\do\o\do\p\do\q\do\r\do\s\do\t%
  \do\u\do\v\do\w\do\x\do\y\do\z\do\A\do\B\do\C\do\D%
  \do\E\do\F\do\G\do\H\do\I\do\J\do\K\do\L\do\M\do\N%
  \do\O\do\P\do\Q\do\R\do\S\do\T\do\U\do\V\do\W\do\X%
  \do\Y\do\Z}

\usepackage{soulutf8}
\usepackage[normalem]{ulem}
\usepackage{xcolor}

\usepackage{enumitem}
\setlist{nolistsep}
\usepackage{microtype}

\iffalse
    \usepackage[final]{proofing}
  \renewcommand\hl[1]{{#1}}  
   {\draftnote{\red{#2}}}
   \newcommand\redHL[1]{}
  \newcommand\todo[1]{}
  \newcommand\barre[1]{}

  \newcommand{\Djame}[1]{}
  \newcommand{\ghazi}[1]{}

\newcommand{\gfcmt}[1]{}

\newcommand{\jlcmt}[1]{}

\newcommand{\dscmt}[1]{}

\else  

\usepackage[draft]{proofing}

\newcommand{\gfcmt}[1]{\textcolor{orange}{#1}}

\newcommand{\jlcmt}[1]{\textcolor{red}{#1}}

\newcommand{\dscmt}[1]{\textcolor{brown}{#1}}

\newcommand{\ghazi}[1]{\textbf{\textcolor{orange}{\hl{Ale: #1}}}}

\newcommand\red[1]{{\color{red}{\bf #1}}}
\newcommand{\barre}[1]{
{\textcolor{red}{{\sout{#1}}}}}

 \newcommand\redHL[1]{\red{\hl{#1}}}
\let\olddraftnote\draftnote
\renewcommand\draftnote[1]{\olddraftnote{\red{#1}}}

\fi

%% file: main.bbl
\begin{thebibliography}{56}
\providecommand{\natexlab}[1]{#1}
\providecommand{\url}[1]{\texttt{#1}}
\expandafter\ifx\csname urlstyle\endcsname\relax
  \providecommand{\doi}[1]{doi: #1}\else
  \providecommand{\doi}{doi: \begingroup \urlstyle{rm}\Url}\fi

\bibitem[Bahdanau et~al.(2015)Bahdanau, Cho, and
  Bengio]{Bahdanau2015NeuralTranslate}
Dzmitry Bahdanau, Kyung~Hyun Cho, and Yoshua Bengio.
\newblock {Neural machine translation by jointly learning to align and
  translate}.
\newblock \emph{3rd International Conference on Learning Representations, ICLR
  2015 - Conference Track Proceedings}, pp.\  1--15, 2015.

\bibitem[Bao et~al.(2020)Bao, Zhou, Huang, Li, Mou, Vechtomova, Dai, and
  Chen]{Bao2020}
Yu~Bao, Hao Zhou, Shujian Huang, Lei Li, Lili Mou, Olga Vechtomova, Xinyu Dai,
  and Jiajun Chen.
\newblock {Generating sentences from disentangled syntactic and semantic
  spaces}.
\newblock \emph{ACL 2019 - 57th Annual Meeting of the Association for
  Computational Linguistics, Proceedings of the Conference}, pp.\  6008--6019,
  2020.
\newblock \doi{10.18653/v1/p19-1602}.

\bibitem[Behjati \& Henderson(2021)Behjati and Henderson]{behjati2021inducing}
Melika Behjati and James Henderson.
\newblock Inducing meaningful units from character sequences with slot
  attention, 2021.

\bibitem[Bowman et~al.(2015)Bowman, Angeli, Potts, and
  Manning]{bowman-etal-2015-large}
Samuel~R. Bowman, Gabor Angeli, Christopher Potts, and Christopher~D. Manning.
\newblock A large annotated corpus for learning natural language inference.
\newblock In \emph{Proceedings of the 2015 Conference on Empirical Methods in
  Natural Language Processing}, pp.\  632--642, Lisbon, Portugal, September
  2015. Association for Computational Linguistics.
\newblock \doi{10.18653/v1/D15-1075}.
\newblock URL \url{https://www.aclweb.org/anthology/D15-1075}.

\bibitem[Bowman et~al.(2016)Bowman, Vilnis, Vinyals, Dai, Jozefowicz, and
  Bengio]{Bowman2016GeneratingSpace}
Samuel~R Bowman, Luke Vilnis, Oriol Vinyals, Andrew~M Dai, Rafal Jozefowicz,
  and Samy Bengio.
\newblock {Generating sentences from a continuous space}.
\newblock \emph{CoNLL 2016 - 20th SIGNLL Conference on Computational Natural
  Language Learning, Proceedings}, pp.\  10--21, 2016.
\newblock \doi{10.18653/v1/k16-1002}.
\newblock URL \url{http://arxiv.org/abs/1511.06349}.

\bibitem[Chen et~al.(2019)Chen, Tang, Wiseman, and
  Gimpel]{Chen2019ARepresentations}
Mingda Chen, Qingming Tang, Sam Wiseman, and Kevin Gimpel.
\newblock {A multi-task approach for disentangling syntax and semantics in
  sentence representations}.
\newblock \emph{NAACL HLT 2019 - 2019 Conference of the North American Chapter
  of the Association for Computational Linguistics: Human Language Technologies
  - Proceedings of the Conference}, 1:\penalty0 2453--2464, 2019.
\newblock \doi{10.18653/v1/n19-1254}.
\newblock URL \url{http://arxiv.org/abs/1904.01173}.

\bibitem[Chen et~al.(2018)Chen, Li, Grosse, and Duvenaud]{Chen2018c}
Tian~Qi Chen, Xuechen Li, Roger Grosse, and David Duvenaud.
\newblock {Isolating sources of disentanglement in variational autoencoders}.
\newblock In \emph{6th International Conference on Learning Representations,
  ICLR 2018 - Workshop Track Proceedings}, 2018.

\bibitem[Cheng et~al.(2020)Cheng, Min, Shen, Malon, Zhang, Li, and
  Carin]{Cheng2020ImprovingGuidance}
Pengyu Cheng, Martin~Renqiang Min, Dinghan Shen, Christopher Malon, Yizhe
  Zhang, Yitong Li, and Lawrence Carin.
\newblock {Improving Disentangled Text Representation Learning with
  Information-Theoretic Guidance}.
\newblock In \emph{Proceedings of the 58th Annual Meeting of the Association
  for Computational Linguistics}, pp.\  7530--7541, 2020.
\newblock \doi{10.18653/v1/2020.acl-main.673}.

\bibitem[Cho et~al.(2014)Cho, van Merrienboer, Gulcehre, Bahdanau, Bougares,
  Schwenk, and Bengio]{cho2014learning}
Kyunghyun Cho, Bart van Merrienboer, Caglar Gulcehre, Dzmitry Bahdanau, Fethi
  Bougares, Holger Schwenk, and Yoshua Bengio.
\newblock Learning phrase representations using rnn encoder-decoder for
  statistical machine translation, 2014.

\bibitem[Clark et~al.(2019)Clark, Khandelwal, Levy, and
  Manning]{Clark2019WhatAttentionb}
Kevin Clark, Urvashi Khandelwal, Omer Levy, and Christopher~D Manning.
\newblock {What Does BERT Look at? An Analysis of BERT’s Attention}.
\newblock In \emph{BlackBoxNLP@ACL}, 2019.
\newblock \doi{10.18653/v1/w19-4828}.
\newblock URL \url{http://arxiv.org/abs/1906.04341}.

\bibitem[de~Marneffe \& Manning(2008)de~Marneffe and
  Manning]{de-marneffe-manning-2008-stanford}
Marie-Catherine de~Marneffe and Christopher~D. Manning.
\newblock The {S}tanford typed dependencies representation.
\newblock In \emph{Coling 2008: Proceedings of the workshop on Cross-Framework
  and Cross-Domain Parser Evaluation}, pp.\  1--8, Manchester, UK, August 2008.
  Coling 2008 Organizing Committee.
\newblock URL \url{https://aclanthology.org/W08-1301}.

\bibitem[De~Marneffe \& Manning(2008)De~Marneffe and Manning]{de2008stanford}
Marie-Catherine De~Marneffe and Christopher~D Manning.
\newblock Stanford typed dependencies manual.
\newblock Technical report, Technical report, Stanford University, 2008.

\bibitem[Devlin et~al.(2019)Devlin, Chang, Lee, and Toutanova]{Devlin2018}
Jacob Devlin, Ming~Wei Chang, Kenton Lee, and Kristina Toutanova.
\newblock {BERT: Pre-training of deep bidirectional transformers for language
  understanding}.
\newblock \emph{NAACL HLT 2019 - 2019 Conference of the North American Chapter
  of the Association for Computational Linguistics: Human Language Technologies
  - Proceedings of the Conference}, 1:\penalty0 4171--4186, 2019.
\newblock ISSN 0140-525X.
\newblock \doi{arXiv:1811.03600v2}.
\newblock URL \url{http://arxiv.org/abs/1810.04805}.

\bibitem[Dittadi et~al.(2021)Dittadi, Tr{\"a}uble, Locatello, Wuthrich,
  Agrawal, Winther, Bauer, and Sch{\"o}lkopf]{dittadi2021on}
Andrea Dittadi, Frederik Tr{\"a}uble, Francesco Locatello, Manuel Wuthrich,
  Vaibhav Agrawal, Ole Winther, Stefan Bauer, and Bernhard Sch{\"o}lkopf.
\newblock On the transfer of disentangled representations in realistic
  settings.
\newblock In \emph{International Conference on Learning Representations}, 2021.
\newblock URL \url{https://openreview.net/forum?id=8VXvj1QNRl1}.

\bibitem[Du et~al.(2020)Du, Lin, Shen, O'Donnell, Bengio, and
  Zhang]{Du2020ExploitingApproach}
Wenyu Du, Zhouhan Lin, Yikang Shen, Timothy~J O'Donnell, Yoshua Bengio, and Yue
  Zhang.
\newblock {Exploiting Syntactic Structure for Better Language Modeling: A
  Syntactic Distance Approach}.
\newblock In \emph{Proceedings of the 58th Annual Meeting of the Association
  for Computational Linguistics}, pp.\  6611–6628, 2020.
\newblock \doi{10.18653/v1/2020.acl-main.591}.
\newblock URL \url{https://www.aclweb.org/anthology/2020.acl-main.591/}.

\bibitem[Dyer et~al.(2016)Dyer, Kuncoro, Ballesteros, and
  Smith]{Dyer2016RecurrentGrammars}
Chris Dyer, Adhiguna Kuncoro, Miguel Ballesteros, and Noah~A Smith.
\newblock {Recurrent neural network grammars}.
\newblock \emph{2016 Conference of the North American Chapter of the
  Association for Computational Linguistics: Human Language Technologies, NAACL
  HLT 2016 - Proceedings of the Conference}, pp.\  199--209, 2016.
\newblock \doi{10.18653/v1/n16-1024}.

\bibitem[Hewitt \& Manning(2019)Hewitt and
  Manning]{hewitt-manning-2019-structural}
John Hewitt and Christopher~D. Manning.
\newblock {A} structural probe for finding syntax in word representations.
\newblock In \emph{Proceedings of the 2019 Conference of the North {A}merican
  Chapter of the Association for Computational Linguistics: Human Language
  Technologies, Volume 1 (Long and Short Papers)}, pp.\  4129--4138,
  Minneapolis, Minnesota, June 2019. Association for Computational Linguistics.
\newblock \doi{10.18653/v1/N19-1419}.
\newblock URL \url{https://aclanthology.org/N19-1419}.

\bibitem[Higgins et~al.(2017)Higgins, Matthey, Pal, Burgess, Glorot, Botvinick,
  Mohamed, and Lerchner]{Higgins2019-VAE:Framework}
Irina Higgins, Loic Matthey, Arka Pal, Christopher Burgess, Xavier Glorot,
  Matthew Botvinick, Shakir Mohamed, and Alexander Lerchner.
\newblock {B-VAE: Learning basic visual concepts with a constrained variational
  framework}.
\newblock \emph{5th International Conference on Learning Representations, ICLR
  2017 - Conference Track Proceedings}, pp.\  1--22, 2017.

\bibitem[Higgins et~al.(2018)Higgins, Pal, Rusu, Matthey, Burgess, Pritzel,
  Botvinick, Blundell, and Lerchner]{higgins2018darla}
Irina Higgins, Arka Pal, Andrei~A. Rusu, Loic Matthey, Christopher~P Burgess,
  Alexander Pritzel, Matthew Botvinick, Charles Blundell, and Alexander
  Lerchner.
\newblock Darla: Improving zero-shot transfer in reinforcement learning, 2018.

\bibitem[Honnibal \& Montani(2017)Honnibal and Montani]{spacy2}
Matthew Honnibal and Ines Montani.
\newblock {spaCy 2}: Natural language understanding with {B}loom embeddings,
  convolutional neural networks and incremental parsing.
\newblock To appear, 2017.

\bibitem[Hu et~al.(2020)Hu, Gauthier, Qian, Wilcox, and Levy]{Hu2020AModels}
Jennifer Hu, Jon Gauthier, Peng Qian, Ethan Wilcox, and Roger~P Levy.
\newblock {A Systematic assessment of syntactic generalization in neural
  language models}.
\newblock In \emph{Proceedings of the 58th Annual Meeting of the Association
  for Computational Linguistics}, pp.\  1725--1744, 2020.
\newblock \doi{10.18653/v1/2020.acl-main.158}.
\newblock URL \url{https://www.aclweb.org/anthology/2020.acl-main.158}.

\bibitem[Huang et~al.(2021)Huang, Huang, and
  Chang]{huang-etal-2021-disentangling}
James~Y. Huang, Kuan-Hao Huang, and Kai-Wei Chang.
\newblock Disentangling semantics and syntax in sentence embeddings with
  pre-trained language models.
\newblock In \emph{Proceedings of the 2021 Conference of the North American
  Chapter of the Association for Computational Linguistics: Human Language
  Technologies}, pp.\  1372--1379, Online, June 2021. Association for
  Computational Linguistics.
\newblock \doi{10.18653/v1/2021.naacl-main.108}.
\newblock URL \url{https://aclanthology.org/2021.naacl-main.108}.

\bibitem[Huang \& Chang(2021)Huang and Chang]{huang-chang-2021-generating}
Kuan-Hao Huang and Kai-Wei Chang.
\newblock Generating syntactically controlled paraphrases without using
  annotated parallel pairs.
\newblock In \emph{Proceedings of the 16th Conference of the European Chapter
  of the Association for Computational Linguistics: Main Volume}, pp.\
  1022--1033, Online, April 2021. Association for Computational Linguistics.
\newblock URL \url{https://aclanthology.org/2021.eacl-main.88}.

\bibitem[Jawahar et~al.(2019)Jawahar, Sagot, Seddah, Benoˆıt, and
  Sagot]{Jawahar2019WhatLanguageACL2019}
Ganesh Jawahar, Benoît Sagot, Djamé Seddah, Ganesh~Jawahar Benoˆıt, and
  Benoˆıt Sagot.
\newblock {What does BERT learn about the structure of language?(ACL2019)}.
\newblock 2019.
\newblock URL \url{https://hal.inria.fr/hal-02131630}.

\bibitem[John et~al.(2020)John, Mou, Bahuleyan, and
  Vechtomova]{John2020DisentangledTransfer}
Vineet John, Lili Mou, Hareesh Bahuleyan, and Olga Vechtomova.
\newblock {Disentangled representation learning for non-parallel text style
  transfer}.
\newblock In \emph{ACL 2019 - 57th Annual Meeting of the Association for
  Computational Linguistics, Proceedings of the Conference}, pp.\  424--434,
  2020.
\newblock ISBN 9781950737482.
\newblock \doi{10.18653/v1/p19-1041}.

\bibitem[Kim \& Mnih(2018)Kim and Mnih]{Kim2018DisentanglingFactorising}
Hyunjik Kim and Andriy Mnih.
\newblock {Disentangling by factorising}.
\newblock \emph{35th International Conference on Machine Learning, ICML 2018},
  6:\penalty0 4153--4171, 2018.

\bibitem[Kingma \& Ba(2015)Kingma and Ba]{Kingma2015}
Diederik~P Kingma and Jimmy Ba.
\newblock {Adam: A Method for Stochastic Optimization}.
\newblock In \emph{3rd International Conference on Learning Representations,
  ICLR 2015, San Diego, CA, USA, May 7-9, 2015, Conference Track Proceedings
  San Diego, CA, USA, May 7-9, 2015, Conference Track Proceedings}, 2015.
\newblock ISBN 9781450300728.
\newblock \doi{10.1145/1830483.1830503}.
\newblock URL \url{http://arxiv.org/abs/1412.6980}.

\bibitem[Kingma \& Welling(2014)Kingma and
  Welling]{DBLP:journals/corr/KingmaW13}
Diederik~P. Kingma and Max Welling.
\newblock Auto-encoding variational bayes.
\newblock In Yoshua Bengio and Yann LeCun (eds.), \emph{2nd International
  Conference on Learning Representations, {ICLR} 2014, Banff, AB, Canada, April
  14-16, 2014, Conference Track Proceedings}, 2014.
\newblock URL \url{http://arxiv.org/abs/1312.6114}.

\bibitem[Kodner \& Gupta(2020)Kodner and Gupta]{Kodner2020OverestimationModels}
Jordan Kodner and Nitish Gupta.
\newblock {Overestimation of syntactic representation in neural language
  models}.
\newblock In \emph{Proceedings of the 58th Annual Meeting of the Association
  for Computational Linguistics}, pp.\  1757–1762, 2020.
\newblock \doi{10.18653/v1/2020.acl-main.160}.
\newblock URL \url{https://www.aclweb.org/anthology/2020.acl-main.160}.

\bibitem[Kulmizev et~al.(2020)Kulmizev, Ravishankar, Abdou, and
  Nivre]{Kulmizev2020DoFormalisms}
Artur Kulmizev, Vinit Ravishankar, Mostafa Abdou, and Joakim Nivre.
\newblock {Do neural language models show preferences for syntactic
  formalisms?}
\newblock In \emph{Proceedings of the 58th Annual Meeting of the Association
  for Computational Linguistics}, pp.\  4077–4091, 2020.
\newblock \doi{10.18653/v1/2020.acl-main.375}.
\newblock URL \url{https://www.aclweb.org/anthology/2020.acl-main.375%0A}.

\bibitem[Lewis et~al.(2019)Lewis, Liu, Goyal, Ghazvininejad, Mohamed, Levy,
  Stoyanov, and Zettlemoyer]{lewis2019bart}
Mike Lewis, Yinhan Liu, Naman Goyal, Marjan Ghazvininejad, Abdelrahman Mohamed,
  Omer Levy, Ves Stoyanov, and Luke Zettlemoyer.
\newblock Bart: Denoising sequence-to-sequence pre-training for natural
  language generation, translation, and comprehension, 2019.

\bibitem[Li et~al.(2020{\natexlab{a}})Li, Gao, Li, Peng, Li, Zhang, and
  Gao]{li2020optimus}
Chunyuan Li, Xiang Gao, Yuan Li, Baolin Peng, Xiujun Li, Yizhe Zhang, and
  Jianfeng Gao.
\newblock Optimus: Organizing sentences via pre-trained modeling of a latent
  space, 2020{\natexlab{a}}.

\bibitem[Li et~al.(2018)Li, Jia, He, and Liang]{li-etal-2018-delete}
Juncen Li, Robin Jia, He~He, and Percy Liang.
\newblock Delete, retrieve, generate: a simple approach to sentiment and style
  transfer.
\newblock In \emph{Proceedings of the 2018 Conference of the North {A}merican
  Chapter of the Association for Computational Linguistics: Human Language
  Technologies, Volume 1 (Long Papers)}, pp.\  1865--1874, New Orleans,
  Louisiana, June 2018. Association for Computational Linguistics.
\newblock \doi{10.18653/v1/N18-1169}.
\newblock URL \url{https://www.aclweb.org/anthology/N18-1169}.

\bibitem[Li et~al.(2020{\natexlab{b}})Li, Murkute, Gyawali, and
  Wang]{Li2020ProgressiveRepresentations}
Zhiyuan Li, Jaideep~Vitthal Murkute, Prashnna~Kumar Gyawali, and Linwei Wang.
\newblock {Progressive Learning and Disentanglement of Hierarchical
  Representations}.
\newblock \emph{arXiv}, 2 2020{\natexlab{b}}.
\newblock ISSN 23318422.
\newblock URL \url{https://openreview.net/forum?id=SJxpsxrYPS
  http://arxiv.org/abs/2002.10549}.

\bibitem[Liu et~al.(2019)Liu, Gardner, Belinkov, Peters, and
  Smith]{liu-etal-2019-linguistic}
Nelson~F. Liu, Matt Gardner, Yonatan Belinkov, Matthew~E. Peters, and Noah~A.
  Smith.
\newblock Linguistic knowledge and transferability of contextual
  representations.
\newblock In \emph{Proceedings of the 2019 Conference of the North {A}merican
  Chapter of the Association for Computational Linguistics: Human Language
  Technologies, Volume 1 (Long and Short Papers)}, pp.\  1073--1094,
  Minneapolis, Minnesota, June 2019. Association for Computational Linguistics.
\newblock \doi{10.18653/v1/N19-1112}.
\newblock URL \url{https://aclanthology.org/N19-1112}.

\bibitem[Locatello et~al.(2020)Locatello, Weissenborn, Unterthiner, Mahendran,
  Heigold, Uszkoreit, Dosovitskiy, and Kipf]{locatello2020objectcentric}
Francesco Locatello, Dirk Weissenborn, Thomas Unterthiner, Aravindh Mahendran,
  Georg Heigold, Jakob Uszkoreit, Alexey Dosovitskiy, and Thomas Kipf.
\newblock Object-centric learning with slot attention, 2020.

\bibitem[Luong et~al.(2015)Luong, Pham, and
  Manning]{Luong2015EffectiveTranslation}
Minh~Thang Luong, Hieu Pham, and Christopher~D Manning.
\newblock {Effective approaches to attention-based neural machine translation}.
\newblock \emph{Conference Proceedings - EMNLP 2015: Conference on Empirical
  Methods in Natural Language Processing}, pp.\  1412--1421, 2015.
\newblock \doi{10.18653/v1/d15-1166}.

\bibitem[Marvin \& Linzen(2020)Marvin and Linzen]{Marvin2020TargetedModels}
Rebecca Marvin and Tal Linzen.
\newblock {Targeted syntactic evaluation of language models}.
\newblock \emph{Proceedings of the 2018 Conference on Empirical Methods in
  Natural Language Processing, EMNLP 2018}, pp.\  1192--1202, 2020.
\newblock \doi{10.18653/v1/d18-1151}.

\bibitem[Nivre et~al.(2016)Nivre, de~Marneffe, Ginter, Goldberg, Haji{\v{c}},
  Manning, McDonald, Petrov, Pyysalo, Silveira, Tsarfaty, and
  Zeman]{nivre-etal-2016-universal}
Joakim Nivre, Marie-Catherine de~Marneffe, Filip Ginter, Yoav Goldberg, Jan
  Haji{\v{c}}, Christopher~D. Manning, Ryan McDonald, Slav Petrov, Sampo
  Pyysalo, Natalia Silveira, Reut Tsarfaty, and Daniel Zeman.
\newblock {U}niversal {D}ependencies v1: A multilingual treebank collection.
\newblock In \emph{Proceedings of the Tenth International Conference on
  Language Resources and Evaluation ({LREC}'16)}, pp.\  1659--1666,
  Portoro{\v{z}}, Slovenia, May 2016. European Language Resources Association
  (ELRA).
\newblock URL \url{https://aclanthology.org/L16-1262}.

\bibitem[Palmer et~al.(2005)Palmer, Gildea, and
  Kingsbury]{palmer2005proposition}
Martha Palmer, Daniel Gildea, and Paul Kingsbury.
\newblock The proposition bank: An annotated corpus of semantic roles.
\newblock \emph{Computational linguistics}, 31\penalty0 (1):\penalty0 71--106,
  2005.

\bibitem[Pimentel et~al.(2020)Pimentel, Valvoda, Hall~Maudslay, Zmigrod,
  Williams, and Cotterell]{pimentel-etal-2020-information}
Tiago Pimentel, Josef Valvoda, Rowan Hall~Maudslay, Ran Zmigrod, Adina
  Williams, and Ryan Cotterell.
\newblock Information-theoretic probing for linguistic structure.
\newblock In \emph{Proceedings of the 58th Annual Meeting of the Association
  for Computational Linguistics}, pp.\  4609--4622, Online, July 2020.
  Association for Computational Linguistics.
\newblock \doi{10.18653/v1/2020.acl-main.420}.
\newblock URL \url{https://aclanthology.org/2020.acl-main.420}.

\bibitem[Raffel et~al.(2020)Raffel, Shazeer, Roberts, Lee, Narang, Matena,
  Zhou, Li, and Liu]{raffel2020exploring}
Colin Raffel, Noam Shazeer, Adam Roberts, Katherine Lee, Sharan Narang, Michael
  Matena, Yanqi Zhou, Wei Li, and Peter~J. Liu.
\newblock Exploring the limits of transfer learning with a unified text-to-text
  transformer, 2020.

\bibitem[Rogers et~al.(2020)Rogers, Kovaleva, and Rumshisky]{Rogers2020AWorks}
Anna Rogers, Olga Kovaleva, and Anna Rumshisky.
\newblock {A Primer in BERTology: What we know about how BERT works}.
\newblock \emph{arXiv}, 2020.

\bibitem[Rolinek et~al.(2019)Rolinek, Zietlow, and
  Martius]{Rolinek2019VariationalAccident}
Michal Rolinek, Dominik Zietlow, and Georg Martius.
\newblock {Variational autoencoders pursue pca directions (by accident)}.
\newblock \emph{Proceedings of the IEEE Computer Society Conference on Computer
  Vision and Pattern Recognition}, 2019-June:\penalty0 12398--12407, 2019.
\newblock ISSN 10636919.
\newblock \doi{10.1109/CVPR.2019.01269}.

\bibitem[Santhanam \& Shaikh(2019)Santhanam and
  Shaikh]{santhanam-shaikh-2019-emotional}
Sashank Santhanam and Samira Shaikh.
\newblock Emotional neural language generation grounded in situational
  contexts.
\newblock In \emph{Proceedings of the 4th Workshop on Computational Creativity
  in Language Generation}, pp.\  22--27, Tokyo, Japan, 29 October--3 November
  2019. Association for Computational Linguistics.
\newblock URL \url{https://www.aclweb.org/anthology/2019.ccnlg-1.3}.

\bibitem[Schmidt et~al.(2020)Schmidt, Mandt, and
  Hofmann]{Schmidt2020AutoregressiveLoops}
Florian Schmidt, Stephan Mandt, and Thomas Hofmann.
\newblock {Autoregressive text generation beyond feedback loops}.
\newblock In \emph{EMNLP-IJCNLP 2019 - 2019 Conference on Empirical Methods in
  Natural Language Processing and 9th International Joint Conference on Natural
  Language Processing, Proceedings of the Conference}, number 2003, pp.\
  3400--3406, 2020.
\newblock ISBN 9781950737901.
\newblock \doi{10.18653/v1/d19-1338}.

\bibitem[Shen et~al.(2019)Shen, Tan, Sordoni, and
  Courville]{Shen2019OrderedNetworks}
Yikang Shen, Shawn Tan, Alessandro Sordoni, and Aaron Courville.
\newblock {Ordered neurons: Integrating tree structures into recurrent neural
  networks}.
\newblock In \emph{7th International Conference on Learning Representations,
  ICLR 2019}, pp.\  1--14, 2019.

\bibitem[Siddhant et~al.(2019)Siddhant, Johnson, Tsai, Arivazhagan, Riesa,
  Bapna, Firat, and Raman]{siddhant2019evaluating}
Aditya Siddhant, Melvin Johnson, Henry Tsai, Naveen Arivazhagan, Jason Riesa,
  Ankur Bapna, Orhan Firat, and Karthik Raman.
\newblock Evaluating the cross-lingual effectiveness of massively multilingual
  neural machine translation, 2019.

\bibitem[Straka(2018)]{straka-2018-udpipe}
Milan Straka.
\newblock {UDP}ipe 2.0 prototype at {C}o{NLL} 2018 {UD} shared task.
\newblock In \emph{Proceedings of the {C}o{NLL} 2018 Shared Task: Multilingual
  Parsing from Raw Text to Universal Dependencies}, pp.\  197--207, Brussels,
  Belgium, October 2018. Association for Computational Linguistics.
\newblock \doi{10.18653/v1/K18-2020}.
\newblock URL \url{https://www.aclweb.org/anthology/K18-2020}.

\bibitem[Tenney et~al.(2020)Tenney, Das, and Pavlick]{Tenney2020BERTPipeline}
Ian Tenney, Dipanjan Das, and Ellie Pavlick.
\newblock {BERT rediscovers the classical NLP pipeline}.
\newblock \emph{ACL 2019 - 57th Annual Meeting of the Association for
  Computational Linguistics, Proceedings of the Conference}, pp.\  4593--4601,
  2020.
\newblock \doi{10.18653/v1/p19-1452}.
\newblock URL \url{http://arxiv.org/abs/1905.05950}.

\bibitem[Vaswani et~al.(2017)Vaswani, Shazeer, Parmar, Uszkoreit, Jones, Gomez,
  Kaiser, and Polosukhin]{Vaswani2017}
Ashish Vaswani, Noam Shazeer, Niki Parmar, Jakob Uszkoreit, Llion Jones,
  Aidan~N Gomez, Lukasz Kaiser, and Illia Polosukhin.
\newblock {Attention Is All You Need}.
\newblock In \emph{NeurIPS}, number Nips, 2017.
\newblock ISBN 1469-8714.
\newblock \doi{10.1017/S0952523813000308}.
\newblock URL \url{http://arxiv.org/abs/1706.03762}.

\bibitem[Weischedel et~al.(2013)Weischedel, Palmer, Marcus, Hovy, Pradhan,
  Ramshaw, Xue, Taylor, Kaufman, Franchini, El-Bachouti, Belvin, and
  Houston]{AB2/MKJJ2R_2013}
Ralph Weischedel, Martha Palmer, Mitchell Marcus, Eduard Hovy, Sameer Pradhan,
  Lance Ramshaw, Nianwen Xue, Ann Taylor, Jeff Kaufman, Michelle Franchini,
  Mohammed El-Bachouti, Robert Belvin, and Ann Houston.
\newblock {OntoNotes Release 5.0}, 2013.
\newblock URL \url{https://hdl.handle.net/11272.1/AB2/MKJJ2R}.

\bibitem[Wu et~al.(2020)Wu, Wang, and Wang]{Wu2020OnBeyond}
Chen Wu, Prince~Zizhuang Wang, and William~Yang Wang.
\newblock {On the Encoder-Decoder Incompatibility in Variational Text Modeling
  and Beyond}.
\newblock 4 2020.
\newblock URL \url{http://arxiv.org/abs/2004.09189}.

\bibitem[Xu et~al.(2020)Xu, Cheung, and Cao]{Xu2020OnSupervision}
Peng Xu, Jackie Chi~Kit Cheung, and Yanshuai Cao.
\newblock {On Variational Learning of Controllable Representations for Text
  without Supervision}.
\newblock \emph{The 37th International Conference on Machine Learning (ICML
  2020)}, 2020.
\newblock URL \url{http://arxiv.org/abs/1905.11975}.

\bibitem[Zhang et~al.(2019)Zhang, Yang, Yuan, Shen, and
  Carin]{Zhang2020Syntax-infusedGeneration}
Xinyuan Zhang, Yi~Yang, Siyang Yuan, Dinghan Shen, and Lawrence Carin.
\newblock {Syntax-Infused Variational Autoencoder for Text Generation}.
\newblock In \emph{Proceedings of the 57th Annual Meeting of the Association
  for Computational Linguistics}, pp.\  2069--2078, Stroudsburg, PA, USA, 6
  2019. Association for Computational Linguistics.
\newblock ISBN 9781950737482.
\newblock \doi{10.18653/v1/P19-1199}.
\newblock URL \url{http://arxiv.org/abs/1906.02181
  https://www.aclweb.org/anthology/P19-1199}.

\bibitem[Zhao et~al.(2017)Zhao, Song, and Ermon]{Zhao2017LearningModels}
Shengjia Zhao, Jiaming Song, and Stefano Ermon.
\newblock {Learning hierarchical features from deep generative models}.
\newblock \emph{34th International Conference on Machine Learning, ICML 2017},
  8:\penalty0 6195--6204, 2017.

\end{thebibliography}
